\documentclass[twocolumn]{article}

\usepackage{graphicx}%
\usepackage{subcaption}
\usepackage{multirow}%
\usepackage{amsmath,amssymb,amsfonts}%
\usepackage{amsthm}%
\usepackage{mathrsfs}%
\usepackage[title]{appendix}%
\usepackage{xcolor}%
\usepackage{textcomp}%
\usepackage{manyfoot}%
\usepackage{booktabs}%
\usepackage{algorithm}%
\usepackage{algpseudocode}%
\usepackage{listings}%
\usepackage{natbib}
\usepackage{authblk}
\usepackage[margin=2cm]{geometry}
\usepackage{hyperref} 
\hypersetup{colorlinks=true,allcolors=[rgb]{0.0,0.4,0.6}} 
\usepackage{cuted}
\newcommand{\vect}[1]{\boldsymbol{#1}}

\newcommand{\equalcontrib}{\textsuperscript{\textdagger}}

\title{Losing dimensions: Geometric memorization in generative diffusion}

\author[1]{Beatrice Achilli\equalcontrib}
\author[1,2]{Enrico Ventura\equalcontrib}
\author[3,4]{Gianluigi Silvestri}
\author[5]{Bao Pham}
\author[6]{Gabriel Raya}
\author[7]{Dmitry Krotov}
\author[1,2]{Carlo Lucibello}
\author[3]{Luca Ambrogioni}

\affil[1]{Computing Sciences Department, Bocconi University, Milan, Italy}

\affil[2]{Bocconi Institute for Data Science and Analytics (BIDSA), Bocconi University, Milan, Italy}

\affil[3]{Donders Institute, Radboud University, Nijmegen, The Netherlands}

\affil[4]{OnePlanet Research Center, Wageningen, The Netherlands}

\affil[5]{Rensselaer Polytechnic Institute, Troy, USA}

\affil[6]{Jheronimus Academy of Data Science (JADS), Tilburg University, s-Hertogenbosch, The Netherlands}

\affil[7]{IBM Research, Cambridge, USA}
\date{}
\begin{document}

\maketitle
\begingroup
\renewcommand\thefootnote{\textdagger}
\footnotetext{These authors contributed equally to this work.}
\endgroup
\begin{strip}
\abstract{Diffusion models power leading generative AI, but when and how they memorize training data—especially on low-dimensional manifolds—remains unclear. We find memorization emerges gradually, not abruptly: as data become scarce, diffusion models experience a smooth collapse where their capacity to vary across independent directions diminishes. Measuring latent dimensionality via the learned score field, we reveal how generative behavior increasingly centers on a few examples while other variations “freeze out”. We propose a geometric memorization theory, showing that salient features collapse first, then finer details, leading to near point-wise replication. This mirrors physical systems condensing into a few low-energy configurations. Our theoretical predictions align with both synthetic and real data, identifying geometric memorization as a distinct phase between generalization and exact copying.}
\end{strip}


Generative diffusion models \citep{sohldickstein2015deep} have achieved spectacular performance in image \citep{ho2020denoising, song2019generative, song2021score} and video \citep{ho2022video, singer2022make, blattmann2023videoldm, videoworldsimulators2024} generation, and currently form the backbone of most state-of-the-art image generation software \citep{betker2023improving, esser2024scaling}. The defining feature of these models is their remarkable ability to generalize over complex, high-dimensional data distributions. However, diffusion models are also known to fully memorize the training set in the low-data regime \citep{somepalli2023diffusion, somepalli2023understanding, kadkhodaie2024generalization}, and in this regime have been shown to be mathematically equivalent to Dense Associative Memory networks \citep{ambrogioni2023search, hoover2023memory, pham2025memorization}, which are modern variants of the celebrated Hopfield model admitting very large memory storage capacity \citep{krotov2016dense}. The ability of these models to memorize has widespread societal implications, as in such cases their adoption could violate existing copyright laws. Therefore, understanding the factors that lead to generalization and memorization has significant practical and theoretical value, driving future developments.

Recent studies have investigated the performance of diffusion models trained on structured data defined over a manifold, in line with the so-called manifold hypothesis \citep{peyre2009manifold, fefferman2016testing}. Recent works \citep{ventura2024spectral, qu2024diffusion, debortoli2023convergence, kadkhodaie2024generalization} analyze generalization aspects, specifically how the target distribution supported by the manifold is reconstructed during diffusion. In particular, \cite{ventura2024spectral} approach the problem from a geometric perspective, characterizing the temporal evolution of the diffusion potential and showing how subspaces of the manifold emerge at different time scales depending on the statistical properties of the data.
Concerning the memorization phenomenon, \cite{pidstrigach2022scorebased} was the first to show that diffusion models are capable of learning low-dimensional structure in $\mathbb{R}^d$, and that this manifold learning capability drives memorization: a model capable of learning $0$-dimensional manifolds can memorize the training data. On the theoretical side, \cite{achilli2025memorization, macris2025analysis} explore how the memory threshold shifts as a function of the global geometry of the manifold and the number of data points in the training set.

While the most natural definition of memorization is about reproducing data points, instead of new samples generated from the same distribution, the underlying phenomenon is more subtle and complex. 
The \textit{reconstructive} memory described in \cite{somepalli2023diffusion} represents the capability of a Diffusion Model to replicate ``objects" that are semantically equivalent to their source object without being pixel-wise identical. \cite{webster2023reproducible} refers to similar instances as template verbatim. 
Moreover, \cite{ross2025geometric} recently proposed the \textit{manifold memorization hypothesis}: the authors evaluate memorization in terms of the relationship between the dimensionalities of the true data manifold and the manifold learned by the model. Their analysis categorizes memorized data into two types: overfitting-driven memorization and memorization driven by the underlying data distribution.

In the works we mentioned above, there is a common idea that memorization can occur separately in different subspaces or directions of the ambient space of the data. However, only a few studies focus on the way data are dynamically memorized, and on how the phenomenology of memorization is affected by an underlying manifold structure in the data.
In fact, when studying memorization, it is natural to ask whether it corresponds to a single transition or whether information is lost gradually until only a single data point can be retrieved. \emph{We conjecture that memorization can be conceived as a progressive loss of degrees of freedom in the diffusive stochastic process}. 
We refer to this progressive memorization phenomenon as \emph{geometric memorization}.

\begin{figure*}[ht!]
    \centering\includegraphics[width=0.8\linewidth]{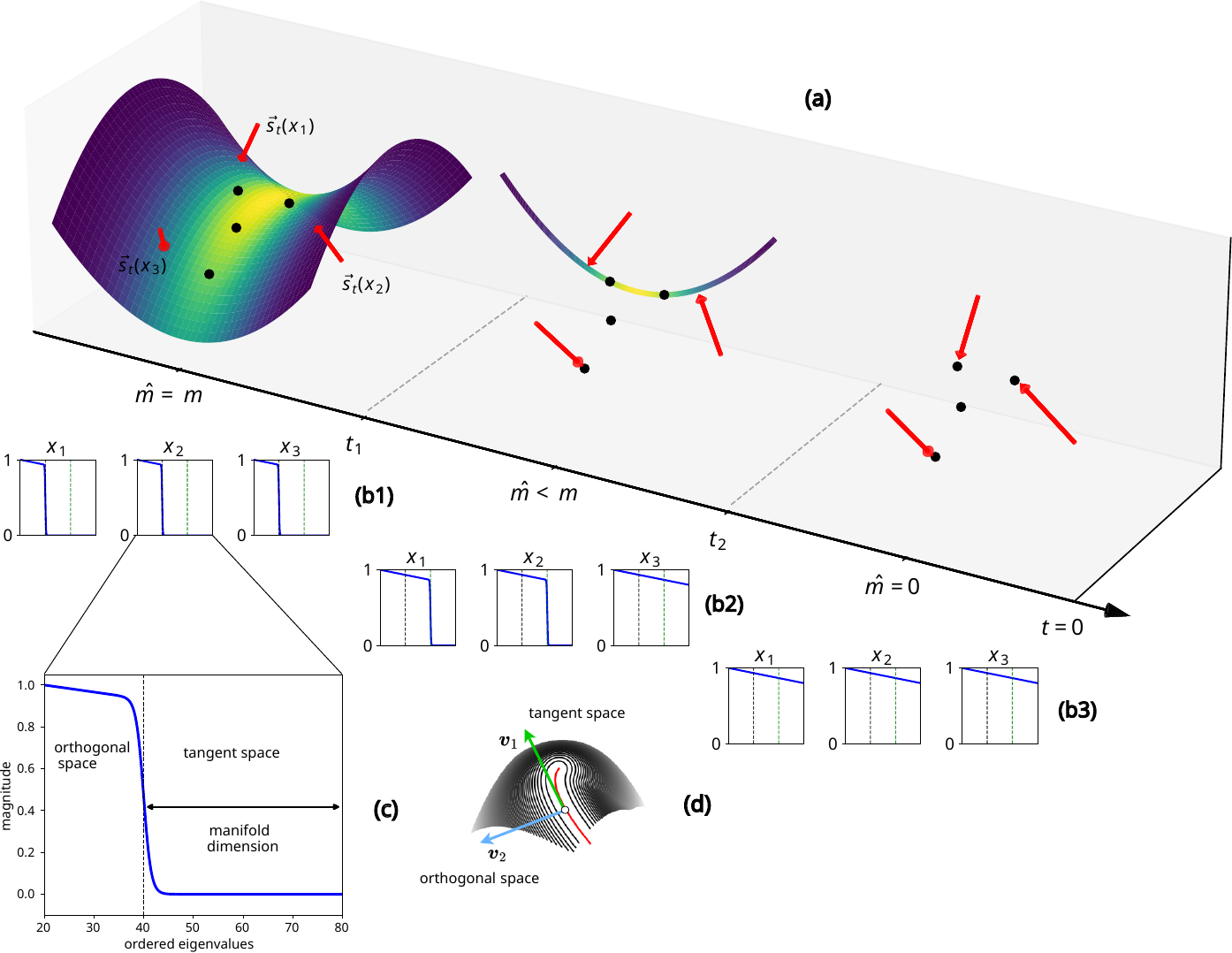}
    
    \caption{Pictorial representation of the geometric memorization phenomenon. Panel (a): data points lie on a $m$-dimensional manifold in the ambient space of dimension $d$, with $m < d$. The target distribution is a 2-variate Gaussian density function with diagonal covariance matrix and variances $\sigma_1^2 > \sigma_2^2$, having the manifold as support. We represent four data points: two being fully aligned with the direction having larger dispersion $\sigma_1^2$, the remaining two being fully aligned with the orthogonal direction. When $t > t_1$ the score function vector field locally projects the sampling process on the manifold (as described in \cite{stanczuk2023your, ventura2024spectral}). We expect a neural network trained on a very large training set to maintain this behavior until very small times $t_1$, ensuring generalization. When $t_1 > t > t_2$ the model starts memorizing fractions of the manifold: points aligned with the larger variance are fully memorized and become point-wise attractors in the vector field; the points aligned with the smaller variance sub-manifold are not yet memorized, and the surrounding score function vectors are orthogonal to the sub-manifold of dimensions $\hat{m} < m$. When $t < t_2 $ all points are attractors in the vector field, and the original $m$-dimensional manifold gets shattered into $0$-dimensional sub-manifolds, i.e. disconnected points in the ambient space. Panels (b): ordered eigenspectrum measured in positions $\vect{x}_1, \vect{x}_2, \vect{x}_3$ according to the Normal Bundle (NB) method described in Section \ref{sec:improved-nb}. When $t > t_1$ spectral gaps opened on the black dashed line, signal an estimated manifold dimension $\hat{m} = m$ all over in space. When $t_1 > t > t_2$ the method applied to $\vect{x_1}, \vect{x}_2$ estimates $\hat{m} < m$ with spectral gaps opened on the green dashed line. The method in $\vect{x}_3$ estimates a $0$-dimensional manifold, i.e. the closest data point is fully memorized, because it lies on the larger variance sub-manifold. When $t < t_2$ spectral gaps estimate $\hat{m} = 0$ all over in space. Panel (c): detailed interpretation of the way the NB method estimates the manifold latent dimension. Panel (d): visualization of the orthogonal and tangent subspaces with respect to a latent manifold.}
    \label{fig:placeholder}
\end{figure*}

To test our hypothesis, we first conduct an experimental exploration with real datasets, which appears to corroborate our conjecture. We then adopt a simple manifold data model and use theoretical analysis to separate memorization events along the directions spanned by the latent manifold.
In the memorization regime, the neural network score of a trained diffusion model approximates the empirical score function. Leveraging the analogy between the empirical score and the Random Energy Model \citep{lucibello2024exponential, biroli2024dynamical, achilli2025memorization}, we develop a theory of geometric memorization based on the analysis of the eigenvalue spectrum of the Jacobian of the \emph{empirical} score function. We find the emergence of spectral gaps that were not predicted by the geometric theory developed in previous works \citep{ventura2024spectral}, providing evidence for a progressive loss of dimensionality due to memorization. The geometric memorization phenomenology, and how this behavior emerges from our analysis is sketched in Figure \ref{fig:placeholder}. 

\textit{Setting} - We will consider a simple variance-exploding forward process, where the data $\vect{x}_0 \sim p_{0}(\vect{x})$ evolves according to the equation
$\text{d}\vect{x}_t= \text{d}\vect{W}_{t}$, where $\text{d}\vect{W}_{t}$ is a standard Brownian motion.  
The \emph{target distribution} $p_{0}(\vect{x})$ is then recovered by reversing the diffusion process \citep{anderson1982reverse}. Starting at $\vect{x}_{T} \sim\mathcal{N}\left(0, T \cdot\mathbf{I}_d\right)$ for some large time $T$, the reverse process SDE is
\begin{equation}
    \text{d}\vect{x}_{\tau}= -\nabla_{\vect{x}}\log p_{\tau}(\vect{x}_{\tau}) \text{d}{\tau}+d\vect{W}_{\tau},
\end{equation}
with ${\tau}$ the backward time corresponding to $\tau = T - t$. We will often describe the reverse process using forward time: the generative process starts at large times $t=T$, and new samples are produced at $t=0$.
The function $s(\vect{x},t)=\nabla_{\vect{x}}\log p_{t}(\vect{x})$ is the so-called score function. 
As for data, we assume the manifold hypothesis, which states that the distribution on natural data (e.g. images, sound recordings) is 
supported on an $m$-dimensional manifold $\mathcal{M}$ embedded in a larger Euclidean ambient space $\mathbb{R}^d$ \citep{peyre2009manifold, fefferman2016testing}. While a probability distribution supported on an $m < d$ manifold $\mathcal{M}$ cannot be expressed using a proper density function, we loosely define such density as 
\begin{equation}
\label{eq:p0}
    p_0(\vect{x}) = \delta_{\mathcal{M}}(\vect{x})\, \rho(\vect{x})~,
\end{equation}
where $\delta_{\mathcal{M}}$ is the Dirac function for the manifold and such that $\int_{\mathbb{R}^d} \delta_{\mathcal{M}}(\vect{x})\bullet d\vect{x} = \int_{\mathcal{M}}\bullet\, d\vect{x}$. We call 
$\rho(\vect{x})$
the \emph{internal density}, that is, the density restricted to the manifold. The density $p_0(\vect{x})$ is zero outside the manifold and diverges on the manifold.

\section{Results}

\subsection{Experimental evidence of Geometric Memorization}\label{sec:exper}
In this Section we provide experimental evidence of geometric memorization in generative diffusion. Such observations have motivated the theory developed in Sec.~\ref{sec:theory}. 

\begin{figure*}[ht!]
    \centering
    \includegraphics[width=0.3\linewidth]{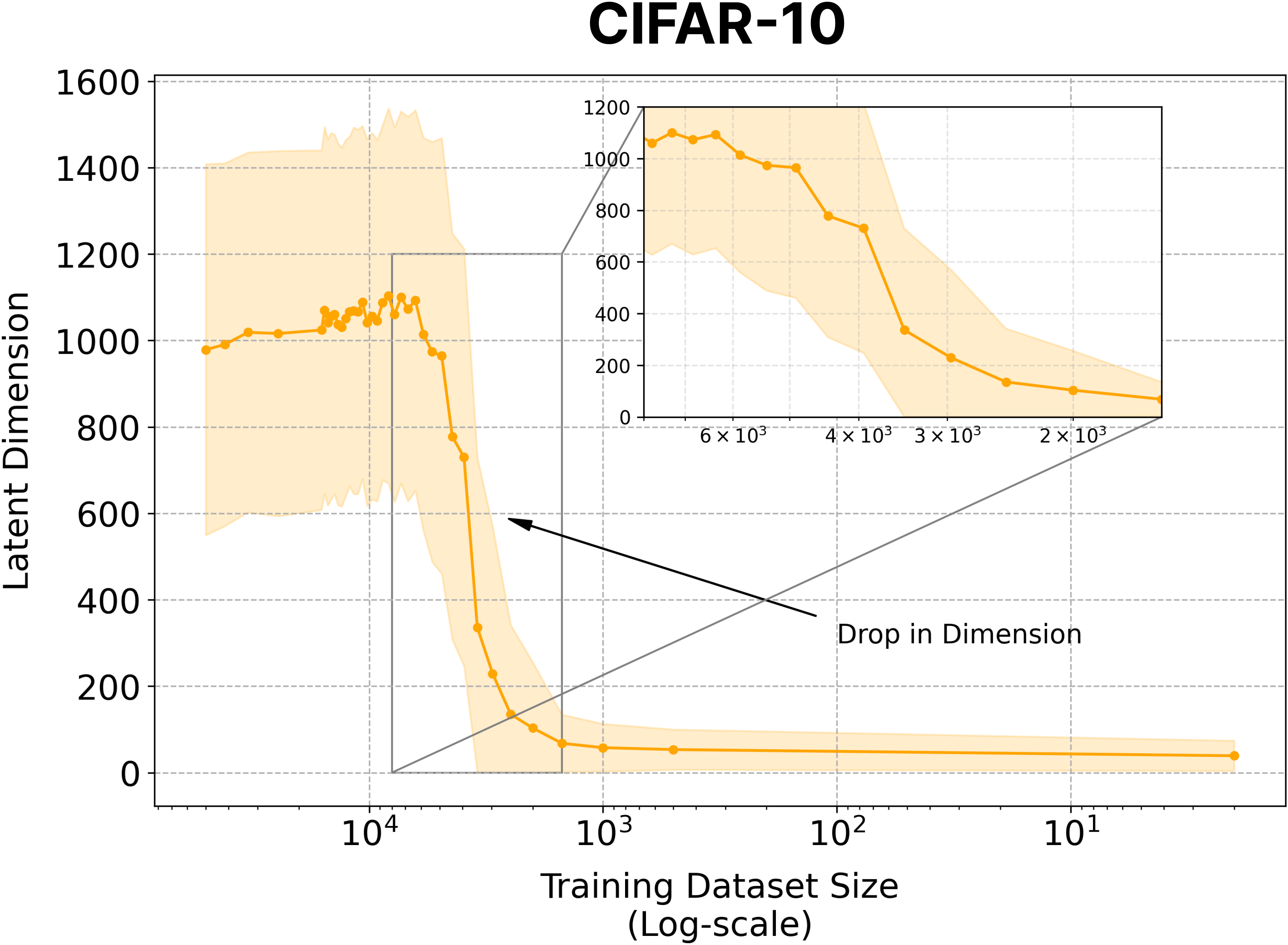}
    \includegraphics[width=0.3\linewidth]{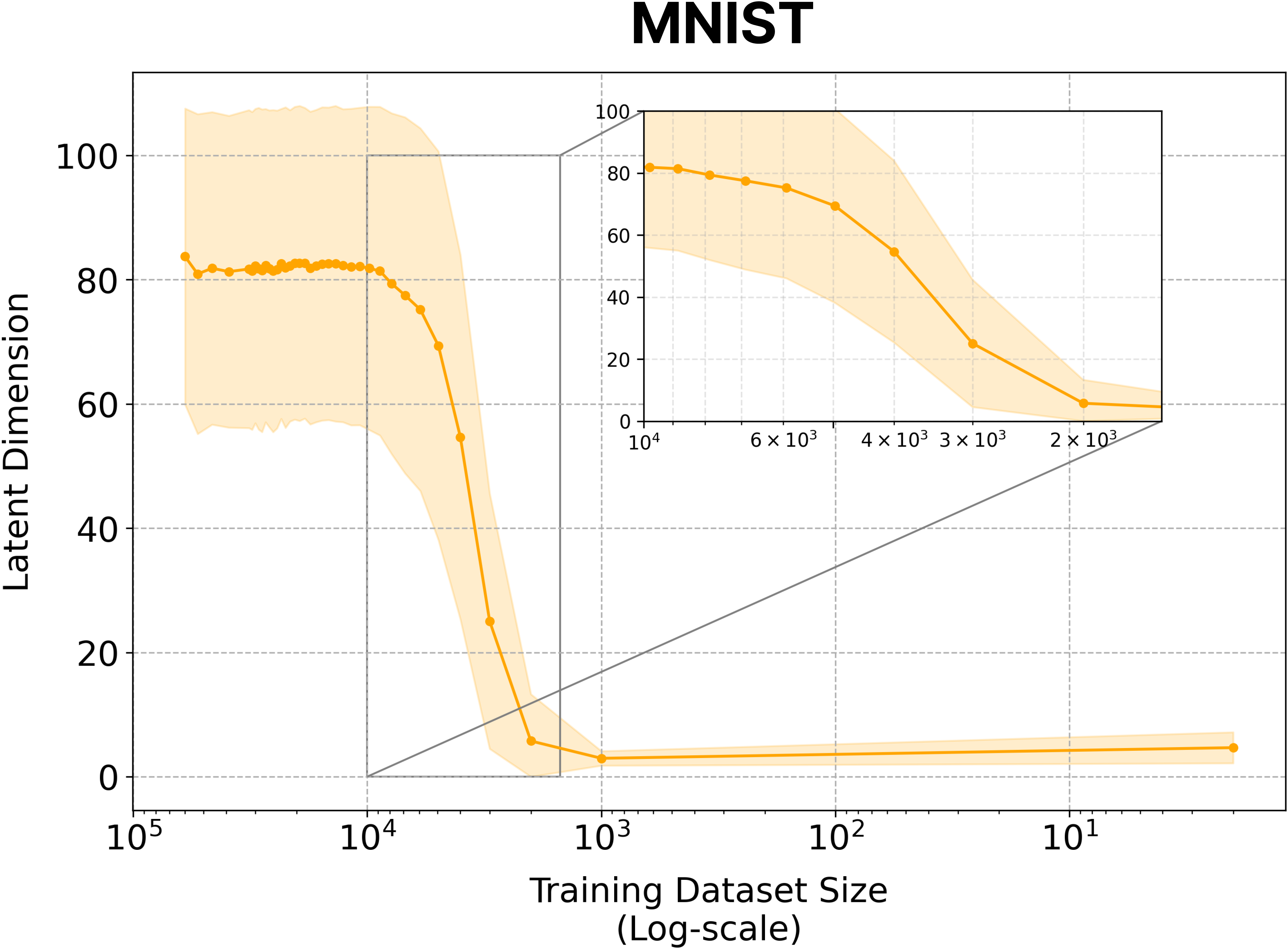}
    \includegraphics[width=0.3\linewidth]{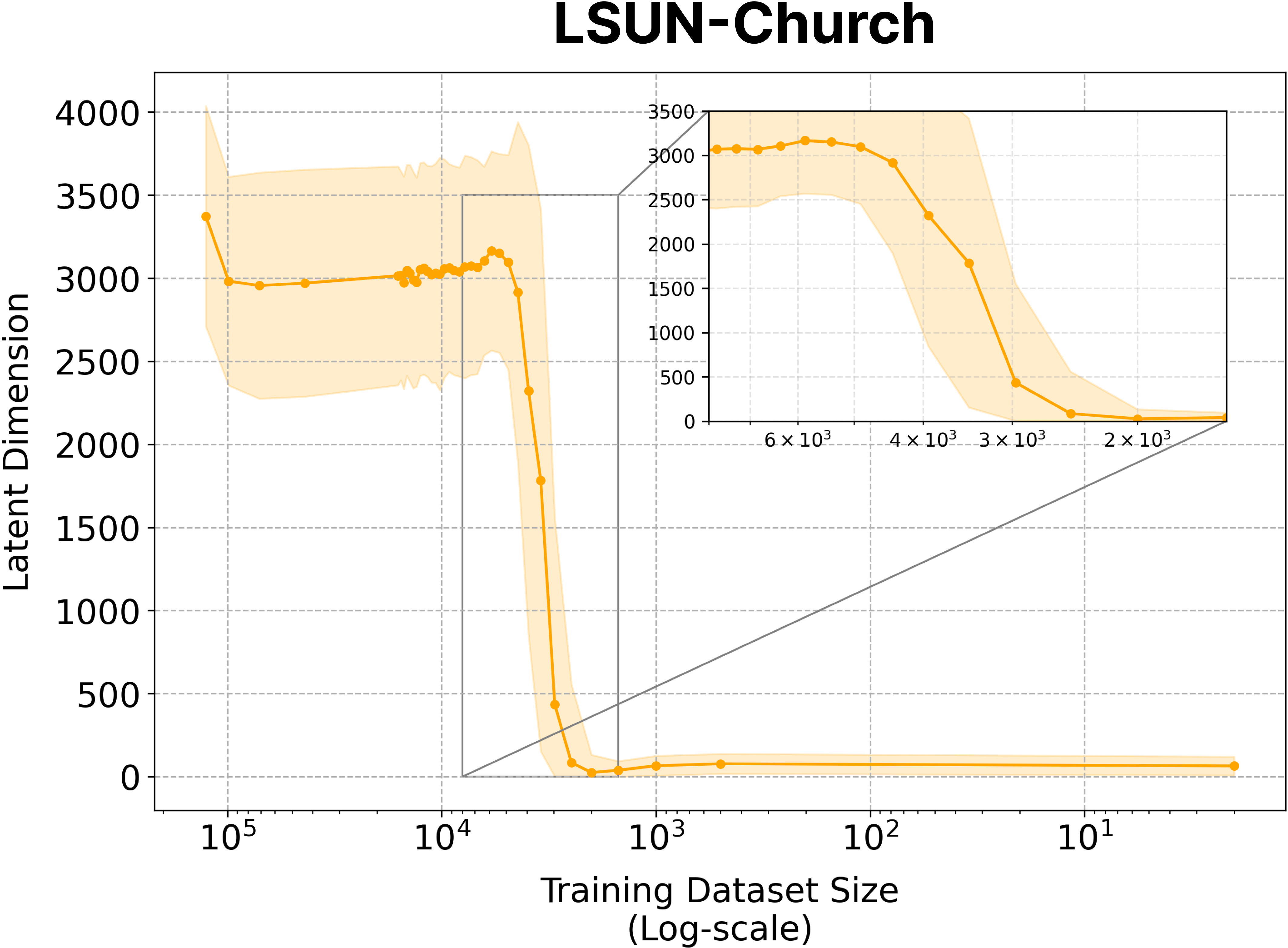}\\
    \includegraphics[width=0.3\linewidth]{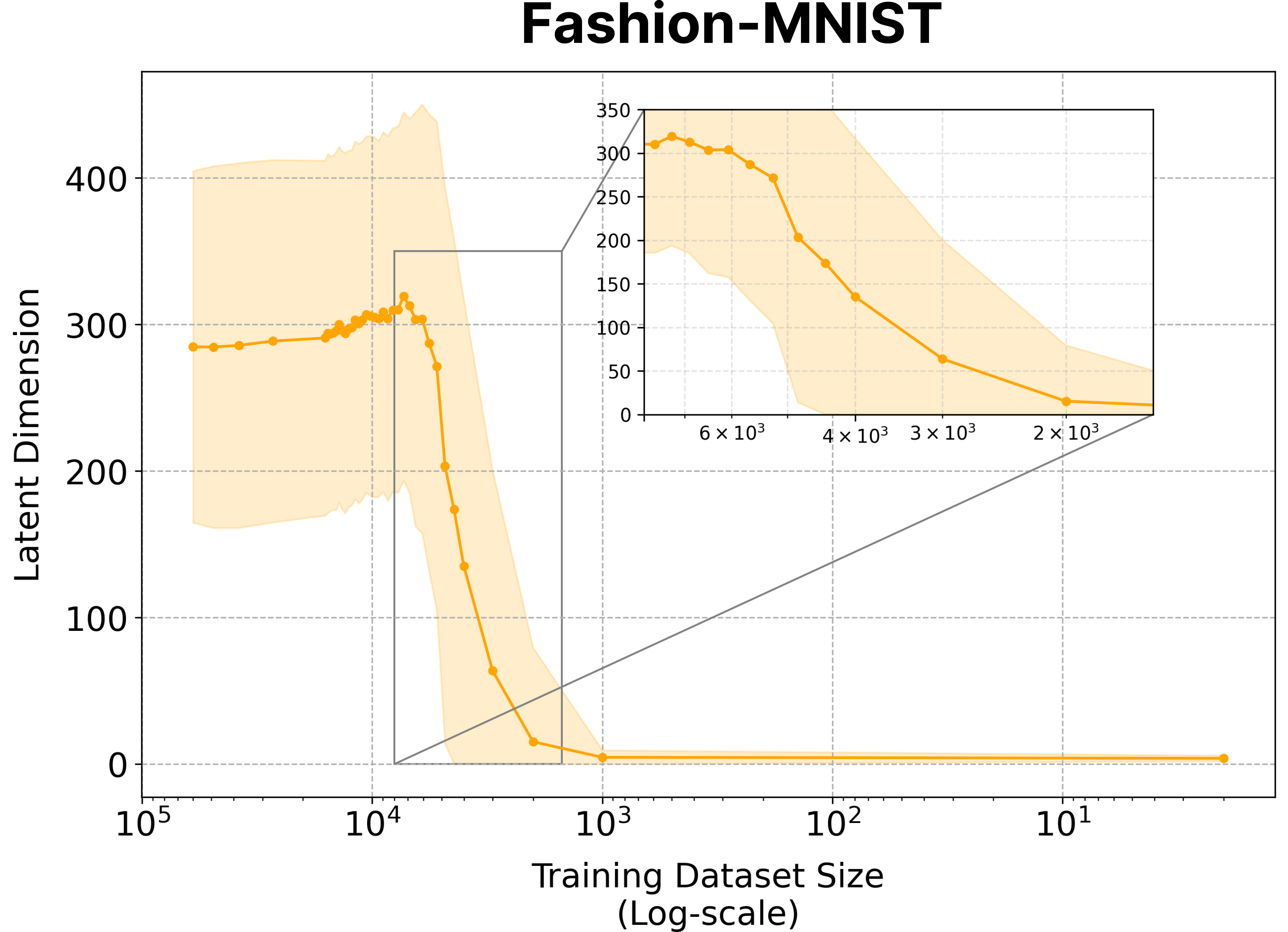}
    \includegraphics[width=0.3\linewidth]{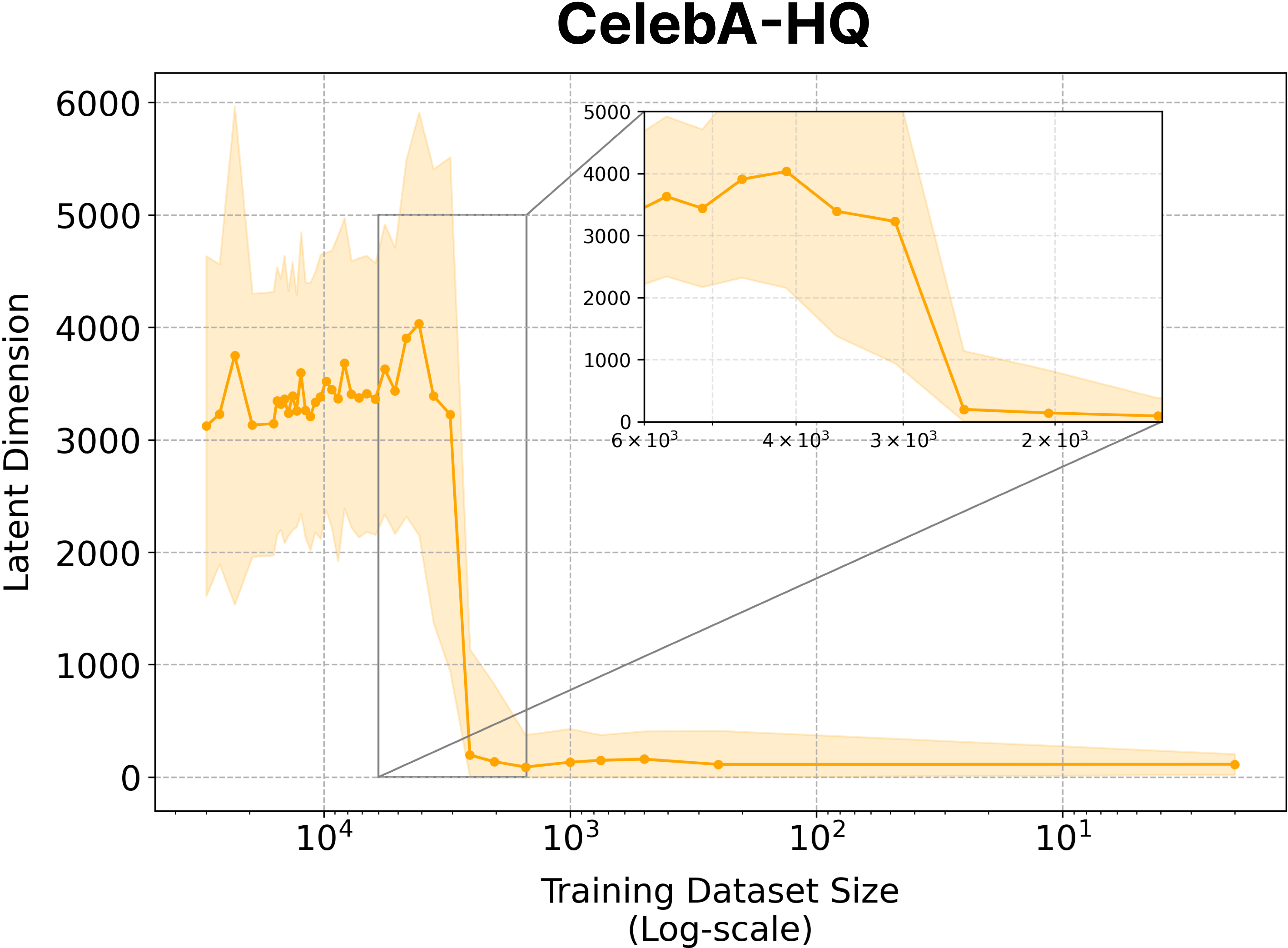}
    
    \caption{Latent manifold dimensionality estimated on deep networks trained on natural image datasets
    at $t = 10^{-5}$.
    The estimated dimensionality tends to smoothly decrease when the dataset size is smaller, suggesting an underlying phenomenon of geometric memorization. Latent dimensionality has been estimated according to Section~\ref{sec:dim}. 
    \label{fig:mnist1}}
\end{figure*}

Let us consider Diffusion Models trained on a series of sub-datasets extracted from MNIST, Cifar10, Fashion-MNIST, CelebA-HQ and LSUN-Churches. 
For each dataset size, we fix a small diffusion time $t = 10^{-5}$, estimate the latent dimensionality around the position $\vect{x}=0$ according to the improved Normal Bundle (NB) analysis described in Section \ref{sec:improved-nb}, and study how this dimensionality varies with dataset size. Full details of the dimensionality estimation procedure and training setup are provided in Section ~\ref{sec:dim}. The average dimensionality as a function of the dataset size is reported in Fig.~\ref{fig:mnist1}. We observe the following progression:
\begin{enumerate}
  \renewcommand{\labelenumi}{\Roman{enumi}.}
\item When the dataset size is very large, i.e. of order $\sim10^4$, the latent dimensionality of the data manifold remains stable: the sample complexity is sufficient, and the network successfully captures the underlying data structure.
\item As the dataset size decreases in the interval $[10^{3}\div 10^4]$, the latent dimensionality gradually declines: the network begins to overfit, but the manifold dimensionality does not collapse abruptly.
\item For even smaller datasets, the latent dimensionality has approached zero, indicating that the network-fitted score function concentrates on individual data points, effectively reducing the manifold to zero-dimensional objects.
\end{enumerate}

\begin{figure*}[ht!]
    \centering\includegraphics[width=0.8\linewidth]{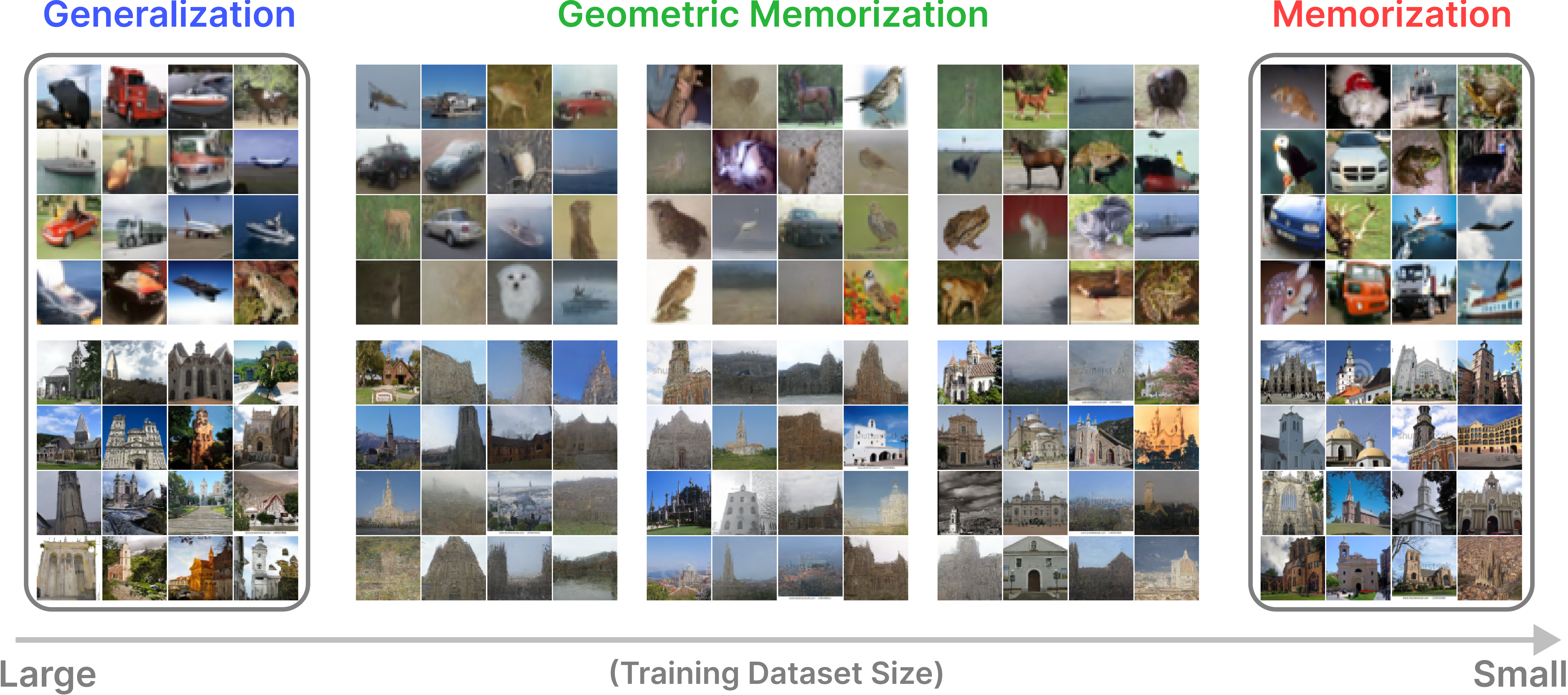}
    \caption{Randomly chosen images that have been generated by diffusion models trained as in Fig. \ref{fig:mnist1}. Training datasets are: Cifar10 (top row) and LSUN-Churches (bottom row). The dataset size decreases from left to right: in the generalization phase, the score function fits the exact one, and images are coherent with a high saturation; during geometric memorization, images look clearly foggy, with a low saturation; when the model memorizes images return to be clear and well saturated since they coincide with existing examples. We conjecture that the reduction in saturation is correlated with the dimensional reduction of the image latent manifold. A related phenomenon was observed in \cite{pham2025memorization} where the intermediate phase was linked to spurious states in Dense Associative Memory models.}
    \label{fig:visual}
\end{figure*}
Previous theoretical and numerical studies of memorization in diffusion models \citep{biroli2024dynamical, achilli2025memorization, lucibello2024exponential, pham2025memorization} have shown that the score function ceases to point towards the data manifold after a characteristic memorization time, which depends on both the properties of the target distribution and the size of the dataset. A natural question is whether, beyond this time, the score field begins orienting towards individual data points abruptly or progressively.

Our results reveal a universal phenomenology across different datasets: memorization unfolds gradually rather than occurring as a sharp transition. In our experiments, the measurement time for dimensionality is fixed for each dataset size. If memorization were abrupt, we would observe a sudden drop in Fig.~\ref{fig:mnist1}, which does not occur. Instead, the latent dimensionality decreases smoothly across a range $[10^{3}\div 10^4]$ in the dataset sizes, suggesting that the diffusion process progressively loses degrees of freedom until it ultimately collapses onto individual data points.

Fig.~\ref{fig:visual} visually represents examples generated by neural networks trained with different dataset sizes, as in Fig.~\ref{fig:mnist1}. For large datasets, the model generalizes, generating new images that are perfectly coherent with the training examples. For intermediate sizes of the dataset, geometric memorization occurs, and generated images experience a drop in saturation: pixel values are more similar to each other, and shapes look foggy. We conjecture that this effect might be correlated with a decrease in the dimensionality of the latent manifold that is connected to a reduction in the number of relevant Fourier modes of the pictures. For small datasets instead, models are fully memorizing, generated images are data points, and saturation is fully restored. We also report that the visual effect of foggy images for intermediate dataset sizes does not evidently appear when we train the model on human faces, as those contained in CelebA-HQ dataset. A deeper analysis of the connection between loss of latent dimensionality and the Fourier decomposition of images will be the object of further investigations. 

\subsection{Theoretical Model of Geometric Memorization}
\label{sec:theory}

In this Section we develop a simple theoretical model inspired by previous descriptions of generative diffusion \citep{biroli2023generative, biroli2024dynamical, achilli2025memorization} that is able to predict, starting from basic elements, the emergence of geometric memorization in  real data. In Section \ref{sec:geom}, we adapt the theory of Random Energy Model (REM) \citep{derrida1981rem} to the geometry of diffusion, and then in Section \ref{sec:eigen} we will employ such results to estimate the eigenspectrum of the Jacobian of the empirical score during memorization. 
\subsubsection{Geometric Memorization Time}\label{sec:geom}
The empirical distribution at time $t$ in the variance exploding framework is
\begin{equation} 
\label{eq:empirical-distribution}
    p_t(\vect{x})=\frac{1}{N\sqrt{(2\pi t)^d}}\sum_{\mu=1}^{N} e^{-\frac{\|\vect{x}-\vect{y}^\mu\|^{2}}{2t}}.
\end{equation}
From the empirical distribution, we can write down the empirical score as
\begin{equation} 
\label{eq:empirical_jac}
    \nabla_{\vect{x}} \log{p_{t}(\vect{x})} = \sum_{\mu=1}^{N} w_{\mu}(\vect{x}, t) \ \frac{\vect{y}^{\mu} - \vect{x}}{t},
\end{equation}
where the weight $w_{\mu}(\vect{x}, t) = p(\vect{y}^\mu | \vect{x};t)/\sum_{\nu=1}^N p(\vect{y}^\nu | \vect{x};t)$ is the posterior probability of the pattern $\vect{y}^{\mu}$ given the noisy state $\vect{x}$, and the possible states are restricted to the empirical set. In the following we will refer to the posterior also as to $p(\vect{x}_0 | \vect{x};t)$. This estimator is consistent, meaning that its bias approaches the true score for $N \rightarrow \infty$. 
The statistical behavior of the empirical score can be analyzed in the large $d$ limit by interpreting Eq.~\eqref{eq:empirical-distribution} as proportional to the partition function of a Random Energy Model (REM) \citep{biroli2024dynamical, lucibello2024exponential}, which offers a simple model of disordered systems. The thermodynamic analysis of generative diffusion models is outlined in \citep{raya2023spontaneous} and \citep{ambrogioni2024thermo}. In summary, each energy level $E_{\mu}$ is associated with a data point $\vect{y}^{\mu}$ in the training set and its energy depends on its Euclidean distance with the current state $\vect{x}_t$ \citep{ambrogioni2024thermo}, with the energy given by 
\begin{equation}
    E_{\mu}(\vect{x}) = -\frac{1}{2} \|\vect{y}^{\mu}\|^{2} + \vect{x} \cdot \vect{y}^{\mu}
\end{equation}
which leads to the partition function
\begin{equation}\label{eq:partition}
    Z(\vect{x}, t) = \sum_{\mu=1}^N e^{-\frac{1}{t} E_{\mu}(\vect{x})}
\end{equation}
where the time parameter $t$ is analogous to the temperature of the system, which can be used to express the weights as a Boltzmann distribution:
$
    w_{\mu}(\vect{x}, t) = \frac{1}{Z (\vect{x}, t)}e^{-\frac{1}{t} E_{\mu}} ~.
$
The number of data points is taken to scales as $N = \exp(\alpha d)$.
Since the empirical score is a Boltzmann average according to Eq.~\eqref{eq:partition}, studying its fluctuation under the random sampling of the data allows us to quantify the deviations from the true score due to memorization effects. In our case, the energy levels are distributed according to
\begin{equation}
    p(E; \vect{x}) = \int_{\mathbb{R}^n} \delta\!\left(E + \frac{1}{2} \|\vect{y}\|^{2} - \vect{x} \cdot \vect{y}\right) \text{d} p_0(\vect{y})
\end{equation}

For small values of $t$ and large dataset sizes, the empirical score can be shown to be self-averaging, meaning that it is insensitive to the specific sampling of the training points, resulting to generalization of the underlying distribution. More formally, from the physical theory of REMs \citep{derrida1981rem}, we know that, at the asymptotic limit of $d \rightarrow \infty$, the statistical system specified by Eq.~\eqref{eq:empirical-distribution} undergoes a phase transition that separates a self-averaging high-temperature regime from a \emph{condensation} regimes where Boltzmann averages depend on a small (i.e. sub-exponential) fraction of energy levels \citep{montanari2009information}. By considering the position $\vect{x}$ as a quenched variable in the problem, we can use the statistical mechanics of REMs to find a positional dependent condensation time. In the more extensive dissertation contained in the Supplementary Material, we show that such position dependent condensation time for linear manifold data is, for a large ratio $\log(N)/d$, equal to
\begin{equation}
    t_c(\vect{x}) = \sqrt{\frac{d}{2\text{log}(N)}\left(\frac{r_{4,\sigma}}{2}+\omega^2(\vect{x})\right)}~,
    \label{eq:tcx}
\end{equation}
which demarcates the diffusion time when the empirical score becomes susceptible to fluctuations introduced by the random sampling of the dataset. When $\log(N)/d$ is not large, the condensation time can be computed through an implicit equation, also reported in the Supplementaries. In Eq.~\eqref{eq:tcx}, the term $r_{4,\sigma}=d^{-1}\sum_{i=1}^d \sigma_i^4$ captures the fluctuations in the norm of the data, while the directional quantity $\omega^2(\vect{x})=d^{-1}\sum_{i=1}^d x_i^2 \sigma_i^2$ is the \emph{variance density} along the direction $\vect{x}$. As we shall see, the balance between these two quantities plays a crucial role in geometric memorization effects. From standard REM calculations \citep{derrida1981rem}, we can express the effective number of data points used to estimate the score at $\vect{x}$ at time $t$ as
\begin{equation}
    \tilde{N}_t(\vect{x}) = \min\!\left(N,\left( 1 - \frac{t}{t_c(\vect{x})}\right)^{-1} \right)~,
\end{equation}
where we introduced the minimum operator heuristically to account for the finite size of the system. 
The exact asymptotic theory is recovered for $d \rightarrow \infty$. 
Note that, since these quantities scale to the leading exponential order, they are therefore neglected quantities that scale sub-exponentially in $N$. 

\subsubsection{Spectral analysis of the Empirical Jacobian}\label{sec:eigen}
The large $d$ analysis outlined in the previous sections gives us a description of the fluctuations in the empirical score as a function of the state $\vect{x}$. The spatial dependency of these fluctuations ultimately depends on the data distribution $p_0(\vect{x})$, which outlines a rich geometric landscape that interacts in a complex way with the spatial variations in the true score $\nabla_{\vect{x}} \log{p_t(\vect{x})}$. 

To study the effect of this spatially non-homogeneous random fluctuations in the spectrum, we start from an approximate formula for the empirical score obtained by restricting the average to only $\tilde{N}_t(\vect{x})$ ``active samples'':
\begin{equation} \label{eq:score-approximation}
    \nabla_{\vect{x}} \log{p_t(\vect{x})} \approx \frac{1}{\tilde{N}_t(\vect{x})} \sum_{\mu \in \text{active samples} } \frac{\vect{y}^\mu - \vect{x}}{t}
\end{equation}
where the ``active samples'' $\vect{y}^{\mu}$ are sampled from the posterior distribution $p(\vect{x}_0 | \vect{x}; t)$. If we consider the data model introduced in~\ref{sec:synth} when the channel function $\Phi$ is the identity, i.e. a simple linear manifold model, Eq.~\eqref{eq:score-approximation} follows a Gaussian distribution, since the posterior is itself Gaussian. 
Let us adopt the simplified hidden manifold data model described in Section \ref{sec:toy-model}, where $m$ canonical directions in the ambient space are characterized with non-zeros variances $\sigma_i^2$, while data are sampled along the rest of the directions with vanishing variances $\sigma_i^2 = 0$. 
If we assume that the fluctuations of the score are uncorrelated for a separation of the order of $\sqrt{t}$, we can quantify the statistical variability of the (smoothed) Jacobian in Eq.~\eqref{eq:smoothed-jacobian} through the following formula
\begin{equation}
\label{eq:empirical-jacobian}
\begin{aligned}
J_{ij}(t) \sim {}& \mathcal{N}\Bigl(
    -\delta_{ij}\bigl(t+\sigma_i^2\bigr)^{-1},\\
&   \frac{\sigma_i^2}{t\bigl(t+\sigma_i^{2}\bigr)}
    \bigl[\phi(t,\vect{0}) + \phi(t,\vect{e}_j\sqrt{t})\bigr]
\Bigr)
\end{aligned}
\end{equation}
where we defined the function 
$
    \phi(t, \vect{x}) = \max \left(1/N, t^{-1} - t_c^{-1}(\vect{x}) \right).
$
On the other hand, in this model gaps can close due to the variance term. We can see this phenomenon qualitatively by considering the singular values spectrum of the expected value of Eq.~\eqref{eq:empirical-jacobian}:
\begin{equation}
   \bar{s}_i = \sqrt{\frac{1}{\left(t+\sigma_i^2 \right)^2} + \sum_{k=1}^d\frac{\sigma_k^2}{t^2\left(t+\sigma_k^{2} \right)^2}\left[\phi(t, \vect{0}) + \phi(t, \vect{e}_i\sqrt{t})\right] ^2 }.
   \label{eq:sibar}
\end{equation}
Remember that we see a gap in the sorted spectrum if there is a large difference between two consecutive sorted singular values $s_k$ and $s_{k+1}$. This gap can disappear mainly for two reasons:
\begin{enumerate}
    \item The first one is that $\phi(t, \vect{e}_k\sqrt{t})$ is larger than $\phi(t, \vect{e}_{k+1}\sqrt{t})$. This case is directional, as it depends on the direction of perturbations $\vect{e}_k$ and $\vect{e}_{k+1}$ and it leads to the selective suppression of a particular subspace.
    \item The second one instead is non-directional: it induces a synchronous suppression of all gaps and leads to complete memorization. It happens if the contribution of these variance terms make the contribution of the expected value negligible.
\end{enumerate}

The first case is the one of our interest. 
For linear Gaussian data, the closing times are determined by the critical time $t_c^{-1}(\vect{x})$, which itself depends on the constant term $r_{4,\sigma}=d^{-1}\sum_{i=1}^d \sigma_i^4$ and on the \emph{directional} term $\omega^2(\vect{x})=d^{-1}\sum_{i=1}^d x_i^2 \sigma_i^2$. This latter term is proportional to the variance along the subspace spanned by $\vect{x}$ and plays a crucial role in determining the differential disappearance of different subspaces at different times. Perhaps counter-intuitively, the subspace spanned by $\vect{e}_k$ is more vulnerable to memorization when $\omega^2(\vect{e}_k)$ is large. Therefore, subspaces that are more prominent in the distribution of the data and that emerge earlier during the diffusion process are also more vulnerable to memorization in the later stages of diffusion. This correspond to the form of feature memorization suggested in \cite{ross2025geometric}.

\begin{figure*}[ht!]
    \centering
        \includegraphics[width=0.24\linewidth]{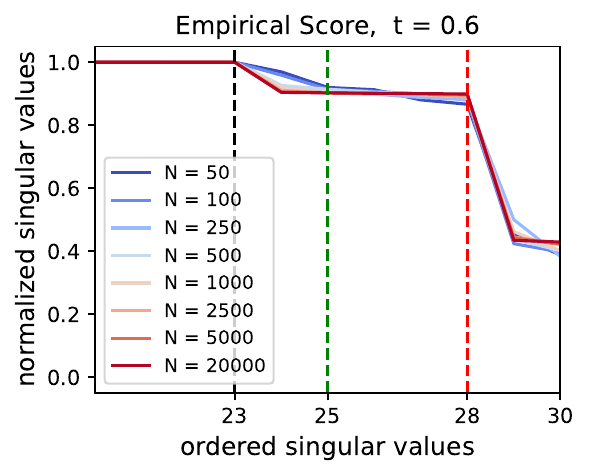}
        \includegraphics[width=0.24\linewidth]{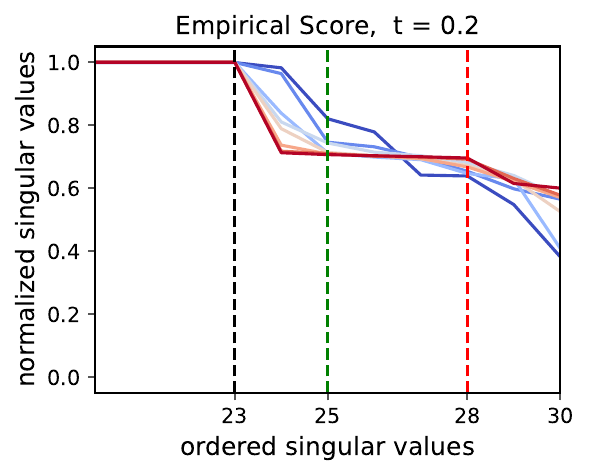}
        \includegraphics[width=0.24\linewidth]{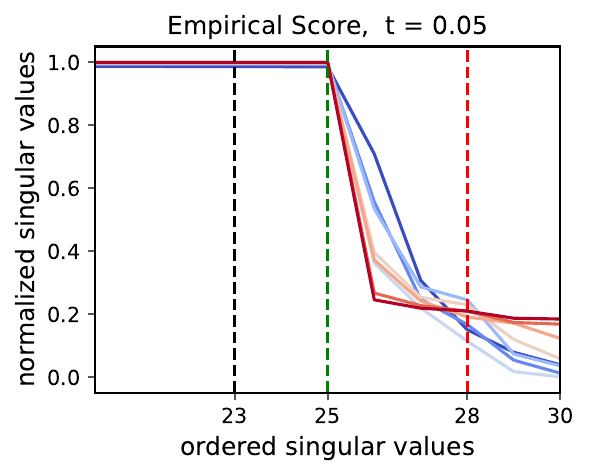}
        \includegraphics[width=0.24\linewidth]{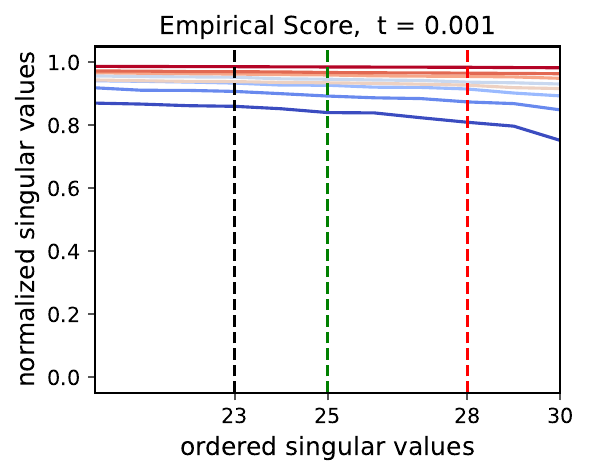} \\
        \includegraphics[width=0.24\linewidth]{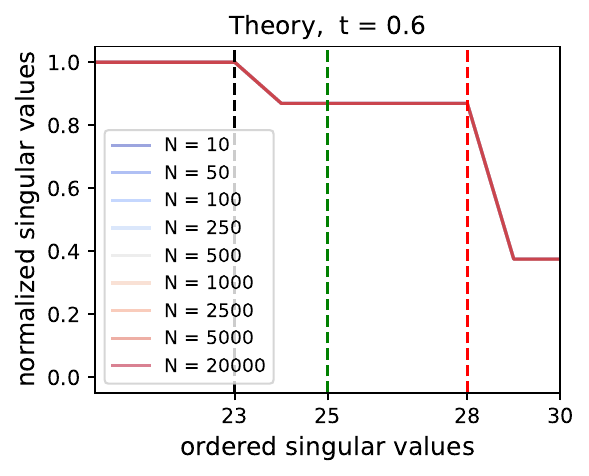}
        \includegraphics[width=0.24\linewidth]{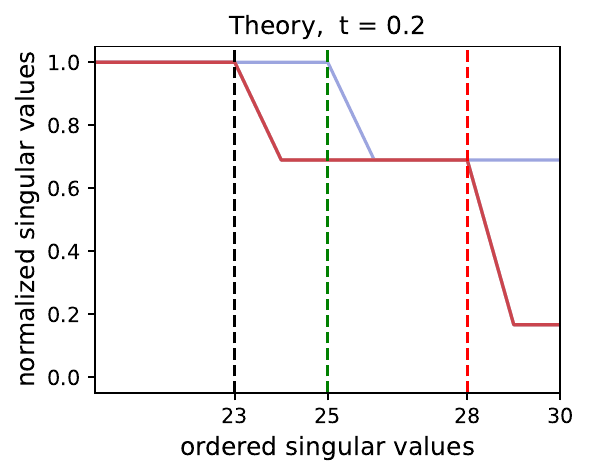}
        \includegraphics[width=0.24\linewidth]{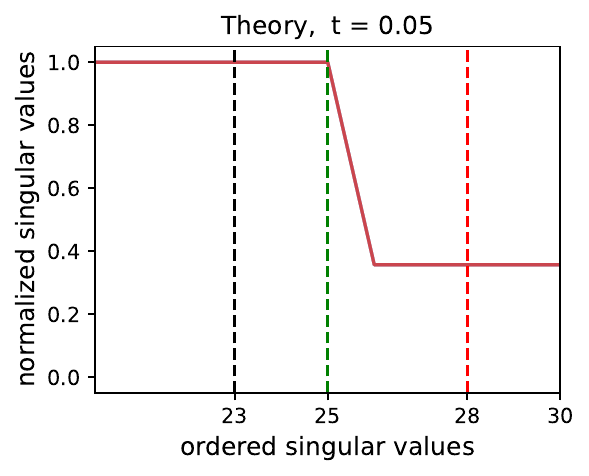}
        \includegraphics[width=0.24\linewidth]{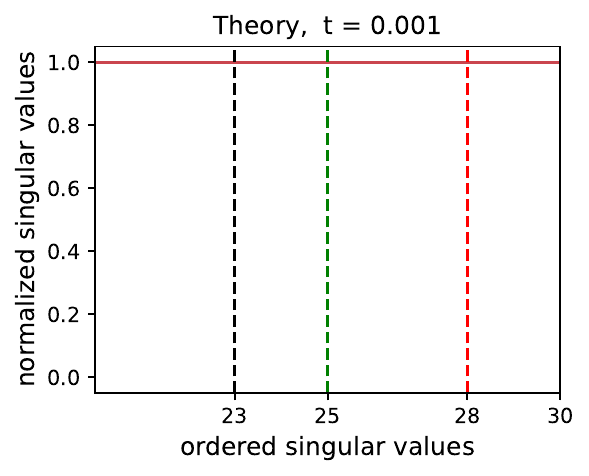}
    
    \caption{Ordered singular values of the Jacobian of the empirical score function in the case of the linear data model, sampled from Eq.~\eqref{eq:empirical-jacobian} (first row) and computed from Eq.~\eqref{eq:sibar} (second row). Choice of the parameters: $d = 30$, $m = 7$ with a subspace associated to a variance $\sigma_1^2 = 1$ of dimension $m_1=2$ and another subspace with variance $\sigma_2^2 = 0.3$ and dimension $m_2 = 5$. Different lines are associated to different sizes of the training set $N$, as reported in the legend. Measures have been averaged over $30$ realizations of the experiment.}
    \label{fig:lin_emp}
\end{figure*}

Figs.~\ref{fig:lin_emp} report the eigenvalues sampled from the distribution in Eq.~\eqref{eq:empirical-jacobian} ordered from the largest to the smallest in magnitude, to reproduce the typical plots obtained by the improved NB method described in Section~\ref{sec:improved-nb}. Figures show drops in magnitude at different eigenvalue numbers, proving the evolution of the local latent dimensionality of the underneath manifold, i.e. a coherent deformation of the score function field in the ambient space. The latent dimensionality decreases in diffusion time, as predicted by the theory. 

On the other hand, Figs.~\ref{fig:lin_emp} also report the ordered eigenvalues yet computed through Eq.~\eqref{eq:sibar}. The curves are consistent with the behavior of the empirical eigenvalues plotted in Figs. and also the experimental curves obtained through the improved NB method. 

\subsection{Comparing Experiments with the Theory}\label{sec:toy-model}
The current Section is devoted to testing the theory developed above through training deep neural networks on the simple data-model described in the Methods (see Section \ref{sec:synth}). We first describe on the network performance in Section \ref{sec:comp1} and then confront it the theory in Section \ref{sec:comp2}. 
\subsubsection{Training Deep Neural Networks \label{sec:comp1}}
\begin{figure*}[t!]
    \centering
    \includegraphics[width=0.8\linewidth]{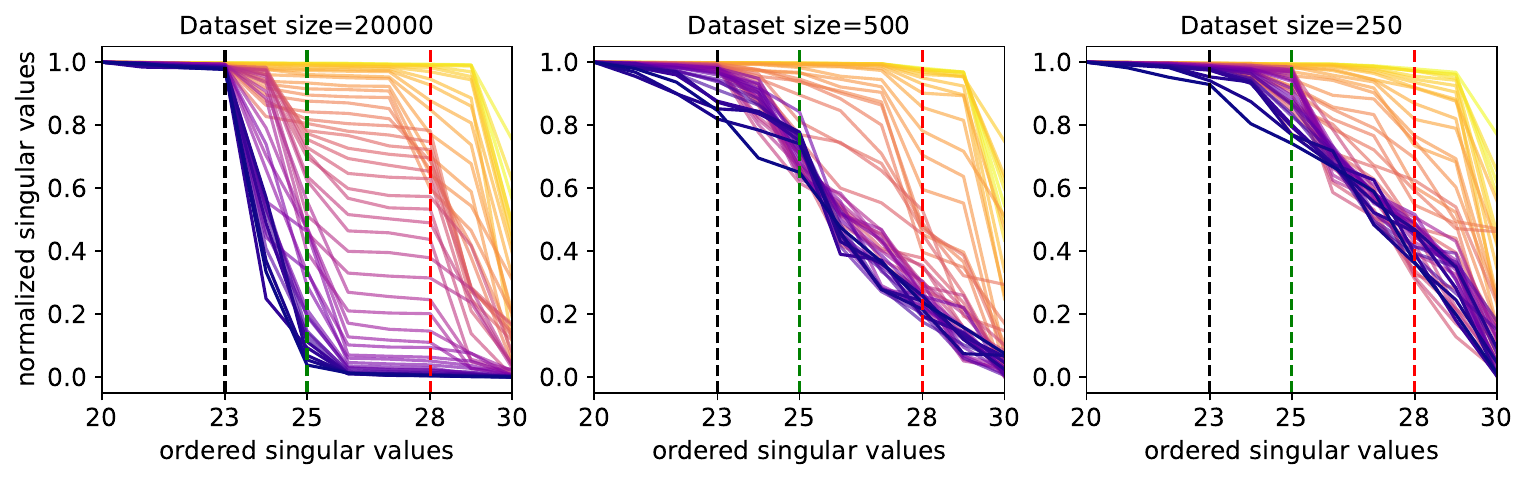}
   \caption{
   The evolution in diffusion time of the ordered singular values of the Jacobian of the score function estimated by a Deep Neural Network trained a linear manifold model. The parameters for the model are $d = 30$, $m = 7$, $\log(N)/d = 0.23$ with subspaces associated to variances $\sigma_1^2 = 1$ and $\sigma_2^2 = 0.3$ with dimensions $m_1=2$ and $m_2 = 5$ respectively. Lighter colors are associated to larger times in the color map.  Different panels have been realized by training a deep network on growing sizes of the training dataset and applying the improved NB method in $\vect{x}=0$. In the Left panel the size of the data-set is sufficiently large to allow the network to detect the true latent dimensionality of the data manifold. In the Central panel the network first starts to reproduce the true dimensionality of the manifold and then enters the geometric memorization phase, opening a gap on the green line, signaling a disruption of such manifold and the estimation of a lower manifold dimension. In the Right panel the network skips the opening of the final gap signaling the true latent manifold dimensions and directly enters the memorization regime, as predicted by our theory. This toy experiment reproduces the  results obtained on real-world data and reported in Section \ref{sec:exper}.}
    \label{fig:nn_spectra}
\end{figure*}

We first perform an experiment on a controlled environment, i.e. by training a deep network on data generated by a simple model. Data have been generated from the synthetic manifold model described in Section \ref{sec:synth}.  
We have considered a linear manifold model with $d=30$, $m=7$, and, in particular, the two subspaces that make the manifold have dimension $m_1=2$ with variance $\sigma_1^2=1.0$ and $m_2=5$ with variance $\sigma_2^2=0.3$. 
We have subsequently trained the neural network on synthetic datasets with an increasing size and applied the improved NB method described in \ref{sec:improved-nb} to estimate the ordered eigenspectrum of the score function Jacobian around the position $\vect{x}=0$ (which we recall to be part of the latent data manifold). The results are reported in Fig.~\ref{fig:nn_spectra}, where the latent dimension of the subspace spanned by the largest variance $\sigma_1^2$ can be estimated by the position of the red dashed line, the dimension of the subspace spanned by the smallest variance $\sigma_2^2$ can be extracted by the green dashed line, and eventually the full latent dimension of the manifold is associated to the black dashed line. 
The left panel in the Figure shows a network trained on a large dataset, which is capable to generalize even at small times. Such model is driven by the true score function that be geometrically reconstructed through the theory developed in \cite{ventura2024spectral}. This network first estimates the smaller latent dimension $m_2$, relative to the sub-manifold spanned by the larger variance $\sigma_2^2$, by opening a gap on the red dashed line, and then opens a last gap on the line relative to the true latent manifold dimension $m$. 
On the other hand, an overfitting model does not manage to geometrically estimate the correct latent dimensionality of the true manifold, because it starts memorizing the data, driven by the empirical score function. This scenario can be appreciated from both central and right panels in Fig. \ref{fig:nn_spectra}.
Specifically, the model represented in the central panel first reproduces the generalizing behavior showed by the left panel, but then opens a gap on the green dashed line at smaller times. This gap is not predicted by a theory based on the true score as the one in \cite{ventura2024spectral}.
On the other hand it can be predicted by a theory based on the empirical score function, and it is a signature of the progressive reduction of the support of the diffusive trajectory that we name geometry memorization. The right panel shows a even more extreme memorization instance, where the model does not manage to open the full manifold dimension gap and instead clearly opens the smaller dimensionality gap. Both the central and right panels would show a flattening of the curves, signaling a zero-dimensional manifold, for very small times, that have not been included in the plot.

\subsubsection{Comparison with the Theory\label{sec:comp2}}
\begin{figure*}[t] 
    \centering
    \includegraphics[width=0.8\linewidth]{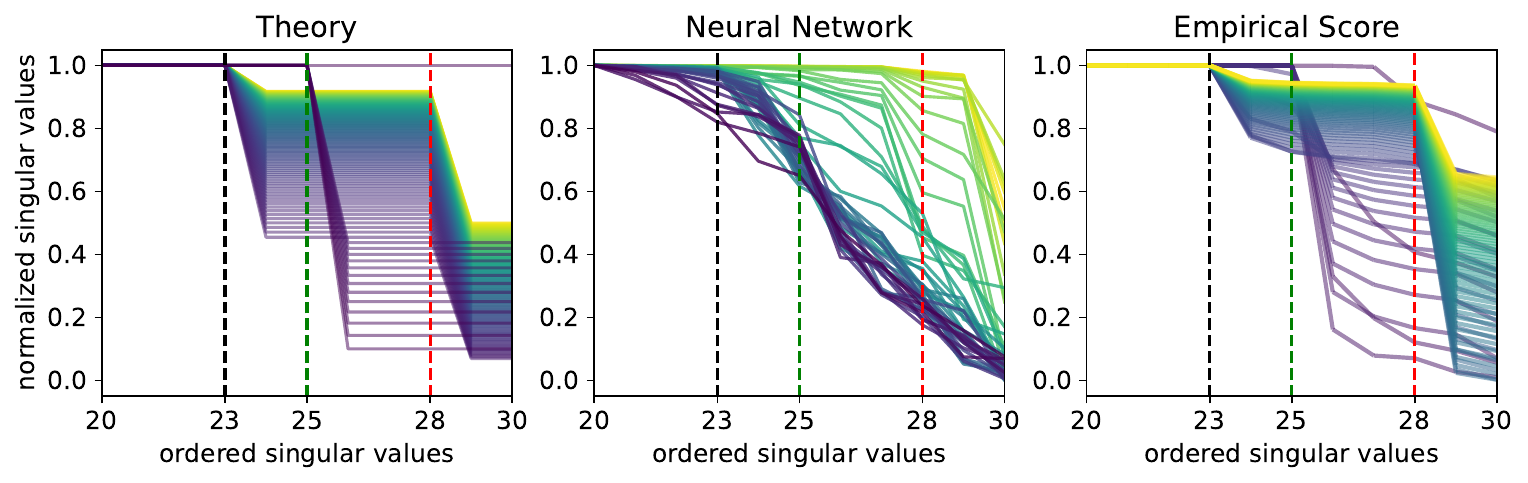}

    \caption{The evolution in time of the ordered singular values of the Jacobian of the score function for a linear manifold model. The parameters for the model are $d = 30$, $m = 7$, $\log(N)/d = 0.23$ with subspaces associated to variances $\sigma_1^2 = 1$ and $\sigma_2^2 = 0.3$ with dimensions $m_1=2$ and $m_2 = 5$ respectively. Lighter colors are associated with larger times in the color map. Left: heuristic theoretical prediction in the memorization phase according to Eq.~\eqref{eq:sibar}. Center: singular values obtained by the improved NB method in $\vect{x}=0$. after training a Neural Network as described in Section \ref{sec:model_details}, over a synthetic dataset of $N = 500$ points. Right: singular values obtained by the numerical measure of the Jacobian of the empirical score function (as described in Section \ref{sec:theory}), evaluated from a synthetic dataset of $N = 10^3$ points.} 
    \label{fig:theoretical-spectra-main-text}
\end{figure*}
Fig.~\ref{fig:theoretical-spectra-main-text} reports the main results from the analysis of geometric memorization on the synthetic manifold model described in Section~\ref{sec:synth} and employed in Section \ref{sec:toy-model}. 
At this point we compare the time evolution of the score function Jacobian computed through three different manners around the position $\vect{x}=0$:
\begin{itemize}
    \item Our theory based on mapping the empirical score on a statistical mechanics model, as described in Section~\ref{sec:theory}.
    \item A neural network model trained according to Section~\ref{sec:model_details}.
    \item A numerical simulation of generative diffusion drive by the empirical score defined in Eq.~\eqref{eq:empirical_jac} in Section \ref{sec:theory}.
\end{itemize}

Starting from large diffusion times, Fig.~\ref{fig:theoretical-spectra-main-text} shows the first gap opening on the red dashed line, signaling that the score function is fitting the $m_1$-dimensional sub-manifold with larger variance $\sigma_1^2$ coherently with the theory presented in \cite{ventura2024spectral}. Subsequently, a gap on the black dashed line appears, suggesting that the score function has reconstructed the full $m$-dimensional manifold. Eventually, this gap closes and leaves room to a new intermediate gap opened on the green dashed-line, associated to a $m_2$-dimensional sub-manifold. The opening of the last gap is a trace of geometric memorization which cannot occur in a true score-driven diffusion model. It suggests that the subspace of higher variance has been memorized, thus it does not count anymore in the manifold dimension. These dimensions have been effectively \textit{lost}, i.e. removed by the subspace spanned by diffusion at small times. Yet it is predicted by our theory based on the spectral analysis of the Jacobian of the empirical score function. This phenomenology is ubiquitous to all the three pictures reported in Fig.~\ref{fig:theoretical-spectra-main-text}. 

\section{Discussion}
In our work, we present experimental evidence and a theoretical analysis of a phenomenon that has so far been rarely observed or discussed in the context of generative diffusion, which we call \textit{Geometric Memorization}. Specifically, we show that data memorization in diffusion models is not a sudden process but a gradual one, deeply intertwined with the structure of the manifold that encodes the correlations and symmetries of a dataset.

Recent studies \citep{ventura2024spectral} have shown that generative models trained on structured data progressively reconstruct both the target distribution and the geometry of the low-dimensional manifold that underlies it. The model first fits the subspaces populated by configurations where features have higher variance and then gradually incorporates dimensions associated with smaller variances. In this way, the model reconstructs large-scale structures of the data before refining the smaller details.
We demonstrate that, when a model overfits the data, it enters a phase in which this manifold is progressively destroyed, following a very similar sequence of steps, but in reverse. The system does not memorize entire data points all at once; instead, it first freezes the features with higher variance and then the finer details, gradually \textit{losing dimensions} until the vector field of the score function becomes fully aligned with the data.
To this end, we introduce a toy data model that can be analyzed within the framework of statistical mechanics. Following the approach of \citep{biroli2024dynamical, ambrogioni2024thermo, achilli2025memorization}, we first map an empirical score-driven diffusion model onto a Random Energy Model, compute the relevant thermodynamic quantities of the system, and then use these to estimate the statistics of the eigenvalues of the Jacobian of the empirical score at small times.

We emphasize that our analysis, ultimately based on the Jacobian of the score function and thus on the local gradient of the diffusion potential, is a static and geometric study of generative diffusion. In contrast, prior works such as \citep{biroli2023generative, biroli2024dynamical} focus on dynamic descriptions, tracking temporal evolution along typical stochastic trajectories. We therefore conjecture that our geometric findings are deeply connected and simultaneously complementary to these dynamic aspects, since the score function governs diffusion, particularly at small times when stochastic noise is minimal.
To sum up, while \citep{qu2024diffusion, ventura2024spectral, ross2025geometric} have highlighted the importance of manifold structure in the generalization behavior of generative diffusion models, our work is the first to explain how the manifold hypothesis influences memorization, suggesting a new way of conceiving data overfitting in these types of machine learning models.  

\section{Methods}

\subsection{Data models}
\label{sec:synth}

For the purpose of our analysis, we will generate data points to have coordinates $y_i^{\mu}\sim \mathcal{N}(0,\sigma_i^2)$ with $\sigma_i^2 > 0$ when $i = 1,..,m$ and $\sigma_i^2 = 0$ when $i = m+1,..,d$. 
This model corresponds to data generation on a linear manifold and is a particular instance of the so-called hidden manifold model proposed by \cite{goldt2020modeling}. The choice of the linear manifold is moved by computational reasons. However, this model is particularly realistic at small diffusion times, where diffusive trajectories are very close to the manifold, as explained in \citep{stanczuk2023your, ventura2024spectral}. \cite{qu2024diffusion} also uses this type of approximation and displays several experimental evidences that corroborate its validity.    


For what concerns our numerical experiments, we employed sub-datasets of various sizes extracted from MNIST, Fashion MNIST, Cifar10, CelebA-HQ and LSUN-Churches. 
Recent works (e.g. \cite{qu2024diffusion, ross2025geometric, brown2023verifying}) give strong evidence that the structure of real data-sets, such as the ones we are using, might be embodied by a latent manifold hidden in the ambient space. Further details about the used datasets are contained in the provided Supplementary material.

\subsection{Training and Model Architecture Details}
\label{sec:model_details}
This Section describes the neural network architectures trained over different natural image data sets.

\begin{table*}[ht!]
\caption{Table displaying both model and training configurations for each dataset. }
    \centering
    \resizebox{0.9\textwidth}{!}{
    \begin{tabular}{ |c|c|c|c|c|c|c| }
     \hline
     \textbf{Dataset}  & \textbf{Image Size} & \textbf{Latent Dim.} & \textbf{Channel Mult.} & \textbf{Param. Count} & \textbf{Batch size} & \textbf{Iterations} \\ 
     \hline 
     Cifar10 & 32 & 128 & (1, 2, 2, 2) & 35.7M & 128 & 500,000 \\ 
     Mnist & 28 & 128 & (1, 2, 2) & 24.5M & 128 & 400,000 \\
     Fashion-Mnist & 28 & 128 & (1, 2, 2) & 24.5M & 128 & 400,000 \\ 
     CelebA-HQ & 64 & 64 & (1, 1, 2, 2, 4, 4) & 27.4M & 64 & 800,000 \\
     Lsun-Church & 64 & 96 & (1, 1, 2, 2, 4, 4) & 61.7M & 64 & 800,000 \\ 
     \hline
    \end{tabular}
    }
    \vspace{2mm}
    \label{tab:models}
\end{table*}
For our toy models, we train a Variance Exploding continuous score model with 2M training steps with batch size 128. We use a Residual Multi Layer Perceptron with hidden size of 128, with two residual blocks. Each block is composed by two linear layers with SiLU activation \citep{Swish}.

For the image models, we follow the diffusion setting in \citep{ho2020denoising}. We kept the variance scheduler, where $\beta_\text{min} = 10^{-4}$ and $\beta_\text{min} = 2 \times 10^{-2}$, the time steps $T = 1000$, and the score model backbone (PixelCNN++ \citep{salimans2017pixelcnn++}) the same. In addition, for each of the datasets, we varied the model's channel multipliers, latent dimension, batch size, and training iterations to account for the complexity of the dataset and our available computing resources; see Table (\ref{tab:models}). For context, we primarily utilized NVIDIA Tesla V100 GPUs with 32 GB of memory for the training of our models.

For what concerns the data sizes that we chose for our experiments, we 
used the pretrained DDPM models in \citep{ho2020denoising, pham2025memorization}: for Cifar10 \citep{cifar10}, Mnist \citep{mnist}, Fashion-Mnist \citep{fmnist}, and Lsun-Church \citep{lsun} datasets. Specifically, these models were trained using $M = 38$ different data split sizes with the goal of finely observing the memorization-to-generalization transition, which also allows us a comprehensive view of the reduction in the manifold size. For each dataset, the smallest model was trained on the training set of $N_1=2$ data points while the largest model was trained on the entire original training set $N_M = N$. 

The selection of these data split sizes relied on spotting a point in which the memorization rate of the model plateaus and another point where its generalization rate increases. Then, linear spacing of 30 points was used between these two points. For example, for CelebA-HQ \citep{liu2015faceattributes} models which we have to trained, these two points are located at 1000 and 16000 data sizes. We used linearly spacing of 30 points between these two points; while for points outside of the transition, we used linearly spacing of 5 points: from 2 to 1000 and 16000 to the full dataset size $N$. For Cifar10, Mnist, and Fashion-Mnist, center-crop and resizing were not used. However, images of the CelebA-HQ and Lsun-Church datasets were both center-cropped and down-sampled to $64 \times 64$ resolution. Finally, we trained our CelebA-HQ models without random flipping and utilized the exponential moving average version of the trained models for our analyses, where we set the decay value to 0.9999 during training. Please refer to Tables (\ref{tab:models})-(\ref{tab:transitions}) for additional details.

\begin{table*}[ht!]
\caption{A table showing the critical points $A$ and $B$ of the memorization-generalization transition for each dataset.}
\label{tab:transitions}
\centering
\begin{tabular*}{\textwidth}{@{\extracolsep\fill}lccc}
\toprule%
& \multicolumn{3}{@{}c@{}}{Training Data Size} \\ \cmidrule{2-4}%
Dataset & Point $A$ & Point $B$ & Total \\
\midrule
Mnist & 4000 & 32000 & 60000 \\ 
Cifar10 & 2000 & 16000 & 50000 \\ 
CelebA-HQ & 2000 & 16000 & 30,000 \\
Lsun-Church & 2000 & 16000 & 126227 \\
Fashion-Mnist &  4000 & 16000 & 60000 \\ 
\bottomrule
\end{tabular*}
\footnotetext{Note: Linear spacing of 30 points between $A$ and $B$ is used to finely characterize the memorization-to-generalization transition according to \cite{pham2025memorization}}
\end{table*}


\subsection{Measuring the intrinsic dimension of the data manifold}\label{sec:method}

In order to probe the geometry of the space where diffusion takes place and its time evolution, we rely on the spectral analysis that we are going to describe in this Section. Through this procedure, it is possible to infer whether the model, at a given position in the ambient space, is constrained to sample along specific directions, signaling the existence of an underlying latent manifold. We underline that this type of study consists of a static analysis of diffusion, that captures the geometric aspects of an underneath evolving manifold, but it does not permit to advance rigorous conclusions in the matter of the sampling dynamics. 

One method used to geometrically estimate the intrinsic dimension of the data manifold is an improved version of Normal Bundle (NB) method used in \citep{stanczuk2023your}: the score function is measured across $d$ orthogonal directions in the vicinity of the manifold and ordered as the columns of a squared matrix $S$; the singular values of the matrix $S$ are computed and collected; the intrinsic dimension of the manifold is given by the $d-\text{ker}(S)$, with the kernel is estimated directly from the spectrum of the singular values of $S$. The algorithm for the improved NB method is described in Section \ref{sec:improved-nb} and the spectral analysis is reported in Section \ref{sec:jacobian_numerical}. 

On the other hand, we propose an alternative estimation method, and we describe it more technically in Section \ref{sec:central}. 
The procedure starts with extracting the singular values of Jacobian of the score function, as performed in the method described above. At this point we order the values by their magnitude and we compute the absolute value of the second derivative of the singular values, selecting the first bigger value with respect to the median multiplied by a threshold factor. We further discard the initial singular value as it tends to be large, resulting in instabilities. The selected singular value signals the formation of a drop in the ordered eigenspectrum, suggesting the beginning of the tangent subspace. Similarly to the previous method, by computing the dimension of the tangent subspace we obtain the best estimate for the latent manifold dimension. We found this method to be more robust than the one proposed in \citep{stanczuk2023your}, especially for high dimensional datasets where there is no sharp drop in the spectrum of the singular values.

\subsubsection{Improved Normal Bundle Method}\label{sec:improved-nb}

Consider a given point $\vect{x}^*$ in the ambient space, and analyze the effect of adding a small perturbation vector $\vect{\varepsilon}$ with magnitude $\|\vect{\varepsilon}\| = \mathcal{O}(\sqrt{t})$:
\begin{equation}
    s(\vect{x}^* + \vect{\varepsilon},t) \approx J(\vect{x}^*, t)~ \vect{\varepsilon}~,
\end{equation}
where $J(\vect{x}^*, t)$ is the Jacobian of the score function. By performing a spectral decomposition of $J(\vect{x}^*, t)$ we can realize that a set of vanishing eigenvalues (or singular values, in experimental instances) must correspond to the tangent space of a latent sub-manifold: the number of vanishing eigenvalues will then correspond to the latent dimension of such manifold. 
This idea, from which we take inspiration, is the basic concept behind the method proposed in Ref. \cite{stanczuk2023your} to estimate the local latent manifold dimension by means of generative diffusion. We refer to this method as the Normal Bundle (NB) method. Given a realization of the diffusive process, one can measure the eigenspectrum in practice, by considering the \textit{smoothed Jacobian matrix} $J(\vect{x}, t)$, whose columns are defined as
 \begin{equation} \label{eq:smoothed-jacobian}
     \vect{J}_j(\vect{x}, t) = \left[s(\vect{x} + \sqrt{t} \vect{e}_j, t) - s(\vect{x}, t)\right]/\sqrt{t}~,
 \end{equation}
 where $\vect{e}_j$ is a vector in an orthonormal basis set. 
 
 The method employed by \cite{stanczuk2023your} relied on sampling a number of random directions that was larger than the dimension of the ambient space, collecting derivatives in a square matrix, and then performing a singular-value decomposition. We have refined such method by sampling $d$ random directions that we subsequently transform into an orthogonal set, to reduce statistical noise and improve the visual quality of the spectra. 
 The defined matrix converges to the exact Jacobian of the score for $t \to 0^+$, while singular-values tend to the true eigenvalues. We will refer to this procedure as to improved NB method. 
 The procedure is reported in Algorithm \ref{alg:sv}.

\begin{algorithm}
\caption{Estimate singular values at $x_0$}
\label{alg:sv}
\begin{algorithmic}[1]
  \Require $s_\theta$ (trained score model), $t_0$ (sampling time), forward process $\mathcal{F}$
  \State Sample $x_0 \sim p_0(x)$ from the data set
  \State $d \gets \mathrm{dim}(x_0)$
  \State $S \gets$ empty matrix
  \For{$i = 1, \dots, d$}
    \State Sample $\epsilon \sim \mathcal{N}(0, I)$
    \State $x_{t_0}^{(i)} \gets \mathcal{F}(x_0, \epsilon, t_0)$ \Comment{ perturbation}
  \EndFor
  \State $(x_{t_0}^{(i)})_{i=1}^d \gets (\tilde{x}_{t_0}^{(i)})_{i=1}^d$ \Comment{orthogonalize the perturbations}
  \For{$i = 1, \dots, d$}
    \State Append $s_\theta(\tilde{x}_{t_0}^{(i)}, t_0)$ as a new column to $S$
  \EndFor
  \State $(s_i)_{i=1}^d,\ (v_i)_{i=1}^d,\ (w_i)_{i=1}^d \gets \mathrm{SVD}(S)$
\end{algorithmic}
\end{algorithm}

In the algorithm, the forward process $\mathcal{F}$ represents forward processes typically used in diffusion models, such as variance exploding and variance preserving noise schedules. Specifically, in the case of the variance exploding schedule, employed by our analysis, the forward process takes the form $\mathcal{F}(x_0, \epsilon, t)=x_0 + t\epsilon$.


\subsubsection{Central difference}\label{sec:central}

We propose a slightly more sophisticated procedure that we call \textit{central difference} method, which is more suitable for natural images datasets, where drops in magnitude might be less visible, both numerical and by human eye \citep{ventura2024spectral}. This technique consists in identifying the singular value where the discrete second derivative of the ordered eigenspectrum is maximum. This point is where the drop in the magnitude of the singular values is the most evident. By subtracting the number associated to this singular value from the full dimensionality of the ambient space, we obtain the latent manifold dimension.
It is reported in Algorithm \ref{alg:sv2}. 


\begin{algorithm}
\caption{Estimate singular values at $x_0$ with central difference}
\label{alg:sv2}
\begin{algorithmic}[1]
  \Require $s_\theta$ (trained score model), $t_0$ (sampling time), forward process $\mathcal{F}$
  \State Sample $x_0 \sim p_0(x)$ from the data set
  \State $d \gets \mathrm{dim}(x_0)$
  \State $S \gets$ empty matrix
  \For{$i = 1, \dots, d$}
    \State Sample $\epsilon \sim \mathcal{N}(0, I)$
    \State $x_{t_0}^{+(i)} \gets \mathcal{F}(x_0, \epsilon, t_0)$ \Comment{right perturbation}
    \State $x_{t_0}^{-(i)} \gets \mathcal{F}(x_0, -\epsilon, t_0)$ \Comment{left perturbation}
  \EndFor
  \State $(x_{t_0}^{+(i)}, x_{t_0}^{-(i)})_{i = 1}^d \gets (\tilde{x}_{t_0}^{+(i)}, \tilde{x}_{t_0}^{-(i)})_{i = 1}^d$ \Comment{orthogonalize the perturbations}
  \For{$i = 1, \dots, d$}
    \State Append $\displaystyle \frac{s_\theta(\tilde{x}_{t_0}^{+(i)}, t_0) - s_\theta(\tilde{x}_{t_0}^{-(i)}, t_0)}{2}$ \\
    as a new column to $S$ \Comment{central difference}
  \EndFor
  \State $(s_i)_{i=1}^d,\ (v_i)_{i=1}^d,\ (w_i)_{i=1}^d \gets \mathrm{SVD}(S)$
\end{algorithmic}
\end{algorithm}

\subsubsection{Spectral Analysis of the estimated Jacobian}\label{sec:jacobian_numerical}

\begin{figure*}[ht!]
    \centering
    \includegraphics[width=0.24\linewidth]{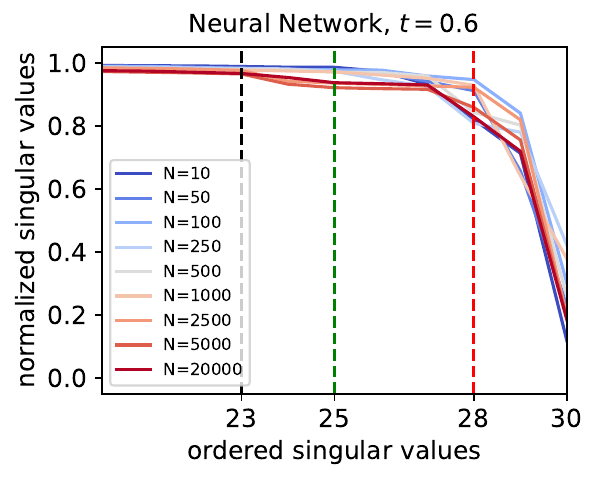}
    \includegraphics[width=0.24\linewidth]{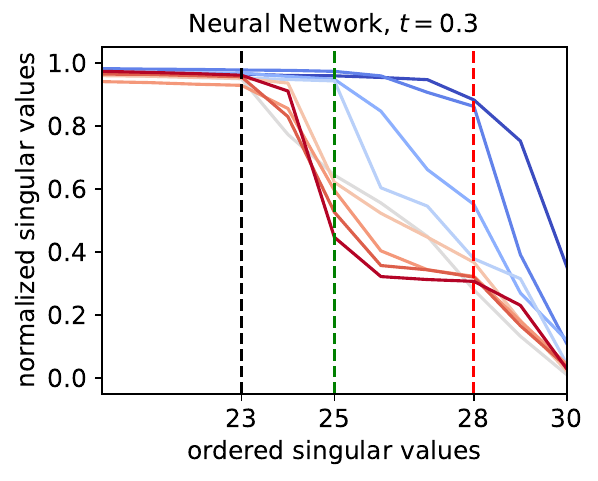} 
    \includegraphics[width=0.24\linewidth]{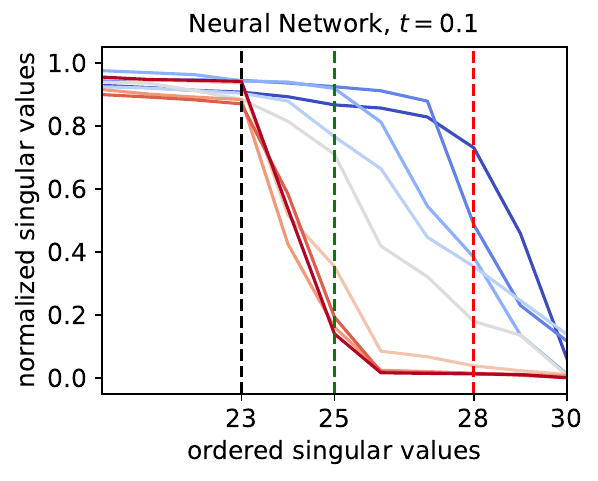}
    \includegraphics[width=0.24\linewidth]{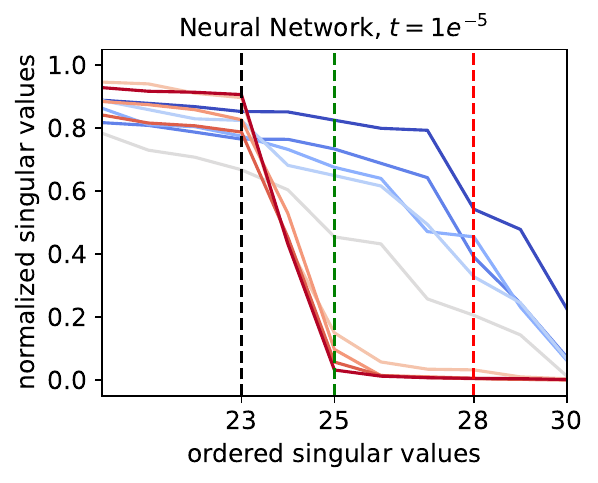}
    \caption{Ordered singular values of the Jacobian of the score estimated by a neural network trained on a linear data model. The parameters for the model are $d = 30$, $m = 7$ with a subspace associated to a variance $\sigma_1^2 = 1$ of dimension $m_1=2$ and another subspace with variance $\sigma_2^2 = 0.3$ and dimension $m_2 = 5$. Different lines are associated to different sizes of the training set $N$, as reported in the legend.}
    \label{fig:lin_score_nn}
\end{figure*}

Once we have extracted the collected the singular values of the Jacobian of the score function throughout Algorithms \ref{alg:sv} and \ref{alg:sv2} we can divide their magnitudes by the highest one and order them from the highest to the slowest. The ordered eigenspectrum plot allows to visualize whether the diffusion model is constrained to sample from a lower dimensional manifold and to infer geometric insights about such manifold. Specifically, drops in magnitude at a certain singular value will display the presence of a gap in the spectrum, which is a signature of a net separation between the tangent and orthogonal subspaces with respect to a manifold. The local dimension of the manifold can be inferred by counting the number of singular values associated to a vanishing magnitude, i.e. relative to the right side of the drop. Figure \ref{fig:lin_score_nn} depicts the time evolution of the ordered eigenspectrum, estimated by the neural network model trained with dataset of different sizes. 

\subsubsection{Estimating the local latent dimension in real datasets}
\label{sec:dim}

\begin{figure*}[ht!]
    \centering
    \includegraphics[width=0.45\linewidth]{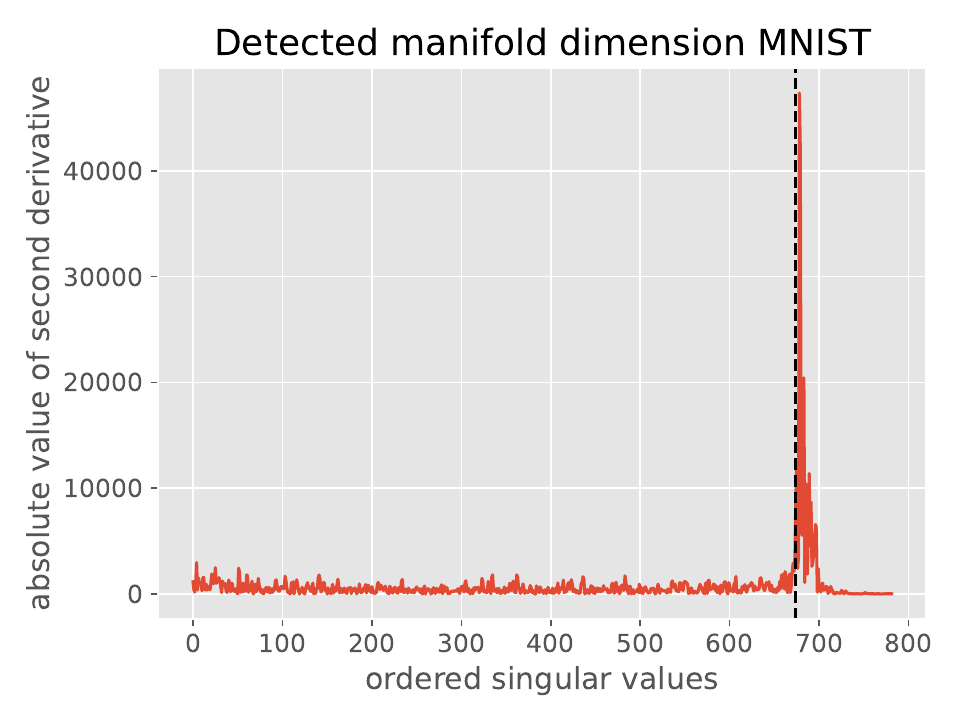}
    \caption{The figure shows the absolute values of the second derivatives computed with algorithm \ref{alg:sv3}, at a small diffusion time when the model is trained with a large batch of the MNIST dataset. The local dimension of the manifold can be estimated by counting the number of singular values on the right of the dashed line.}
    \label{fig:mnistms}
\end{figure*}

In this Section, we report Algorithm~(\ref{alg:sv3}) that we used to find the local latent manifold dimension given the singular values. For Mnist, we used $\Bar{d}=100$, $c=20$; for Cifar10, we used $\Bar{d}=1000$, $c=10$; for CelebA-HQ, we used $\Bar{d}=1500$, $c=10$; and for Lsun-Church, we used $\Bar{d}=2000$, $c=20$. For all these datasets, we employed the same $t_0 = 2$ corresponding to the diffusion index in DDPM. With the exception of CelebA-HQ, which we used Algorithm~(\ref{alg:sv}), we primarily used the central-difference version explained in Algorithm~(\ref{alg:sv2}). We report an example of second derivative used in the method in Fig.~(\ref{fig:mnistms}).
\begin{algorithm}
\caption{Estimate intrinsic manifold dimension at $x_0$}
\label{alg:sv3}
\begin{algorithmic}[1]
  \Require Singular values $(s_i)_{i=1}^d$ from Alg.~\ref{alg:sv} or \ref{alg:sv2}, diffusion time $t$, data dimension $d$, threshold $c$, starting index $\bar{d}$
  \State $d_{\mathrm{svd}}^2 \gets \left|\frac{d^2}{ds^2} s_t[\bar{d}:]\right|$ \Comment{second derivative tail spectrum}
  \State $m \gets \mathrm{median}(d_{\mathrm{svd}}^2)$
  \State $n \gets \#\{i \mid d_{\mathrm{svd},i}^2 > c \cdot m\}$ \Comment{indices above threshold}
  \State $k \gets d - n + \bar{d}$
  \State \Return estimated intrinsic manifold dimension $k$
\end{algorithmic}
\end{algorithm}

\subsection{Theory details}\label{sec:theory_details}
This Section gives the details of the theoretical framework that we have introduced in Section \ref{sec:theory}. Section \ref{app:positional-rem} explains how to use the physics of disordered systems to compute a memorization threshold for the diffusion model; Section \ref{app:jacobian} shows how to use the previous result to analytically estimate the spectrum of the eigenvalues of the Jacobian of the empirical score function, which drives diffusion while the model is memorizing data points.  

\subsubsection{Condensation time for positional REM}\label{app:positional-rem}

For simplicity, we will perform the analysis for coordinate-aligned linear manifolds. Consider $d$-dimensional normally distributed vector-valued data $\vect{y}^\mu$ where each component $y^\mu_k$ follows a centered normal distribution with variance $\sigma^2_k$. In the linear manifold case, number $d-m$ of these variances is equal to zero, meaning that the distribution spans a $m$-dimensional linear manifold. The number of data points are taken to be exponential in the size of the ambient space, i.e. $N = \exp\left(\alpha d\right)$, with $\alpha > 0$. 
Let us take a fixed $\boldsymbol{x}$.
Hence, in the variance exploding framework we have

\begin{align}
    p_{t}(\boldsymbol{x})&=\frac{1}{N\sqrt{2\pi t}^{d}}\sum_{\mu=0}^{N}e^{-\frac{1}{2t}\|\vect{x}-\vect{y}^{\mu}\|^{2}}\\
    &=\frac{1}{N\sqrt{2\pi t}^{d}}e^{-\frac{\|\vect{x}\|^{2}}{2t}}\sum_{\mu=0}^{N}\exp\left(-\frac{1}{2t}\|\vect{y}^{\mu}\|^{2}+\frac{1}{t}\vect{x}\vect{y}^{\mu}\right).
\end{align}
It is useful, at this point, to introduce the Random Energy Model (REM), firstly proposed by physicists \citep{derrida1981rem,montanari2009information}, now imported to computer science to characterize diffusion models \citep{biroli2024dynamical}. The REM consists in a collection of energy levels $\{E_{\mu}\}_{\mu \leq N}$ that interact with an external heat-bath at a temperature $t$. 
The energy levels are random variables generated from a probability density function $p(E|\vect{\theta})$ where $\vect{\theta}$ can be some parameters of the model and source of disorder for the system. 
The thermodynamics of the model shows a condensation phase at a critical temperature $t_c$ that shares similarities with glassy transitions in spin-glass models \citep{parisi}. 
Condensation, in turn, is analogous to memorization in diffusion models. The main thermodynamic quantities, such as the condensation temperature, can be fully recovered starting from the \textit{partition function} of the system, given by
\begin{equation}
\label{eq:part}
    Z_N(\beta) = \sum_{\mu=1}^{N} e^{- \frac{1}{t} E_\mu} .
\end{equation}
We can now map our diffusion model into a REM by redefining 
\begin{equation}
    E_{\mu}(\vect{x}) =  \frac{1}{2}\|\vect{x}-\vect{y}^{\mu}\|^{2}.
\end{equation}
We call this model \textit{positional} REM because the occurrence of condensation will depend on a position in the $d$-dimensional Euclidean space. Standard REM calculations are now performed to compute the free energy of the model and then the condensation time. The moment generating function of the energies is

\begin{align}\label{eq:zeta}
    \zeta(\lambda)&=\lim_{d\to\infty}\frac{1}{d}\log\mathbb{E}_{\vect{y}}e^{-\frac{\lambda}{2t}\|\vect{y}\|^{2}+\frac{\lambda}{t}\vect{x}\vect{y}}\\
    &=\lim_{d\to\infty}\frac{1}{d}\sum_{i=1}^{d}\log\int\frac{dy_{i}}{\sqrt{2\pi\sigma_{i}^{2}}}\exp-\frac{y_{i}^{2}}{2}\left(\frac{1}{\sigma_{i}^{2}}+\frac{\lambda}{t}\right)+\frac{\lambda}{t}x_{i}y_{i}\\
    &=\lim_{d\to\infty}\frac{1}{d}\left[-\frac{1}{2}\sum_{i=1}^{d}\log\left(1+\lambda\frac{\sigma_{i}^{2}}{t}\right)+\frac{\lambda^{2}}{2t^{2}}\sum_{i=1}^{d}\frac{x_{i}^{2}\sigma_{i}^{2}}{1+\lambda\frac{\sigma_{i}^{2}}{t}}\right]
\end{align}
The derivative of the zeta function is
\begin{equation}\label{eq:zeta1}
\begin{aligned}
\zeta'(\lambda)
= \lim_{d\to\infty}\frac{1}{d}\Biggl[
    &-\frac{1}{2t}\sum_{i}\frac{\sigma_i^{2}}{1+\lambda\frac{\sigma_i^{2}}{t}}
     +\frac{\lambda}{t^{2}}\sum_{i}\frac{x_i^{2}\sigma_i^{2}}{1+\lambda\frac{\sigma_i^{2}}{t}}\\
    &-\frac{\lambda^{2}}{2t^{3}}\sum_{i}\frac{x_i^{2}\sigma_i^{4}}
        {\left(1+\lambda\frac{\sigma_i^{2}}{t}\right)^{2}}
\Biggr].
\end{aligned}
\end{equation}
At large times, $\zeta(\lambda)$ and $\zeta'(\lambda)$ become respectively
\begin{equation}
\label{eq:zeta_large}
\zeta(\lambda)=-\frac{\lambda}{2t}r_{2,\sigma}+\frac{\lambda^{2}}{4t^{2}}r_{4,\sigma}+\frac{\lambda^{2}}{2t^{2}}\omega^2(\vect{x}),
\end{equation}

\begin{equation}
\label{eq:zetap_large}
\zeta'(\lambda)=-\frac{\lambda}{2t}r_{2,\sigma}+\frac{\lambda^{2}}{2t^{2}}r_{4,\sigma}+\frac{\lambda^{2}}{t^{2}}\omega^2(\vect{x}).
\end{equation}
Where 
\begin{align}
    r_{2,\sigma} &= \lim_{d\rightarrow\infty}\frac{1}{d}\sum_{i}\sigma_{i}^{2}\\
    r_{4,\sigma} &= \lim_{d\rightarrow\infty}\frac{1}{d}\sum_{i}(\sigma_{i}^{2})^{2}\\
    \omega^2(\vect{x}) &= \lim_{d\rightarrow\infty}\frac{1}{d}\sum_{i}(x_{i})^{2}\sigma_{i}^{2}.
\end{align}
The condition for the condensation of the REM is $\alpha+\zeta(1)-\zeta'(1)=0$. For sufficiently small values of $\alpha$ we obtain larger condensation times. In this case we can employ the expressions in Eqs.~\eqref{eq:zeta_large}, \eqref{eq:zetap_large} to obtain an explicit formula for the condensation time that reads

\begin{equation}
\label{eq:tcx2}
t_{c}(\vect{x})=\sqrt{\frac{\frac{r_{4,\sigma}}{2}+\omega^2(\vect{x})}{2\alpha}}.
\end{equation}

\begin{figure*}[ht!]
    \centering
    \includegraphics[width=0.285\linewidth]{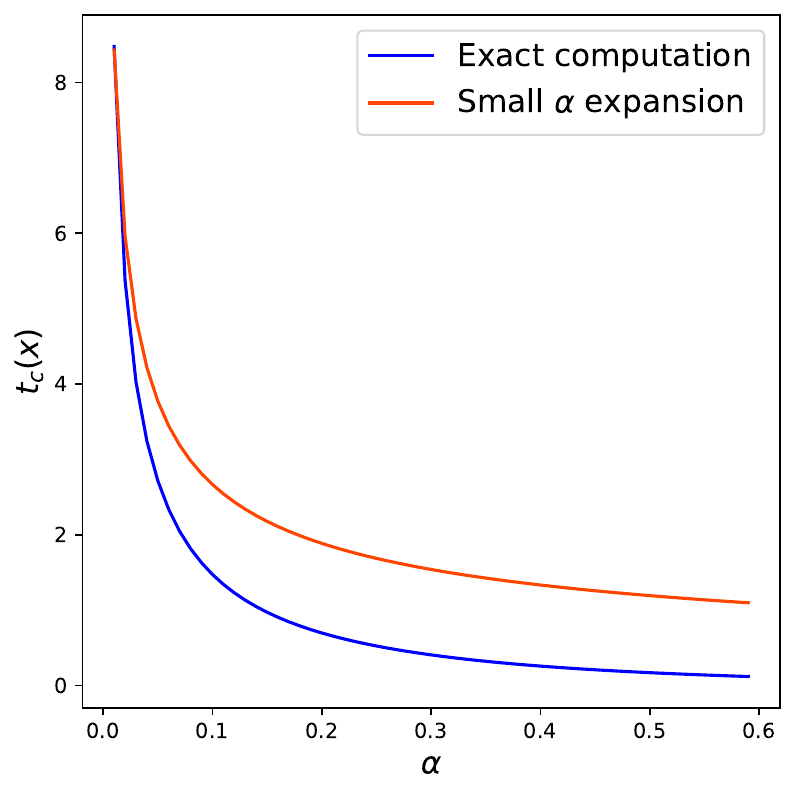}
    \includegraphics[width=0.3\linewidth]{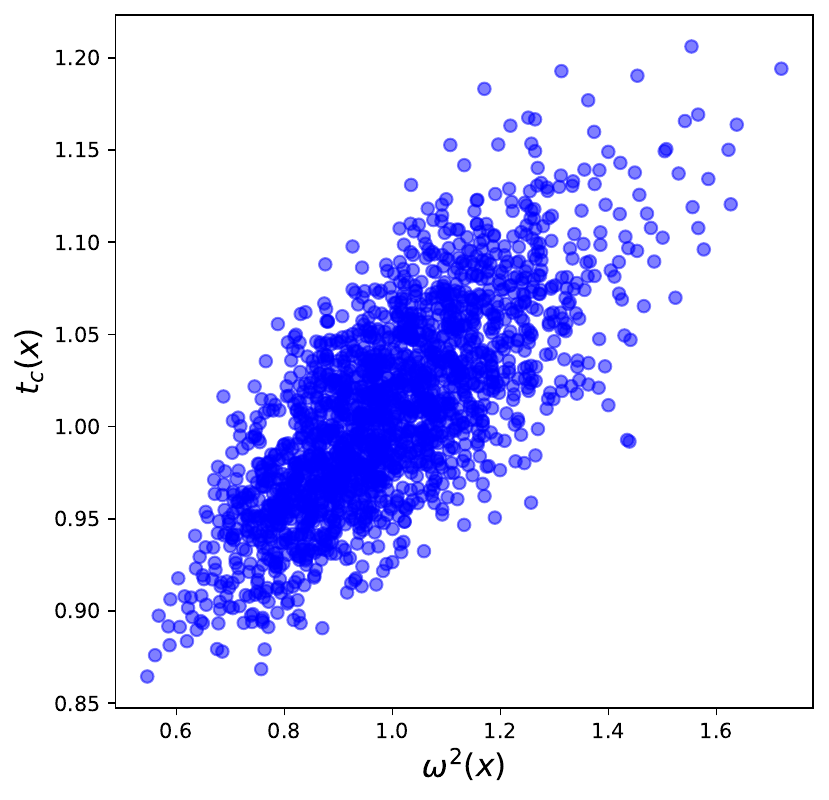}
    
    \caption{Condensation time as a function of the position, computed according to the REM analysis. Left: we have generated one single position $\vect{x}$ in a ambient space of dimension $d = 100$ and one single matrix $F$ of dimensions $100 \times 50$ (with $m = 50$ dimension of the latent space). Both $\vect{x}$ and $F$ are generated according to a Gaussian process with zero mean and unitary variance; we show the comparison between the exact calculation of the positional condensation time and the approximated version that is fully explicit in the directional variance $\omega^2(\vect{x})$. Right: we generate $2000$ random positions $\vect{x}$ around the origin of the ambient space of dimension $d = 100$; the latent space dimension is $m = 50$ and $\alpha = 0.15$; we show the dependence of the exact positional condensation time as a function of $\omega^2(\vect{x})$, showing a qualitatively similar behavior with respect to the approximated expression of $t_c$.}
    \label{fig:tc_comp}
\end{figure*}

As clear from the formula, this time depends on the variance $\omega^2(\vect{x})$ along the direction of $\vect{x}$. This implies that, when $\vect{x}$ is aligned to a linear sub-manifold with higher variance, condensation around this state will happen earlier, leading to a decrease in the estimated commonality of the latent manifold. 
For a better visualization of the meaning of $\omega^2(\vect{x})$, one can consider the specific positions $\vect{x}$ where $x_i = \sqrt{d}$ and $x_{j\neq i} = 0$: in this case $\omega^2(\vect{x}) \approx \sqrt{\frac{\sigma_i^2}{2\alpha}}$ where we used the fact that $r_{4,\sigma} = \mathcal{O}(\frac{m}{d})$ and $m \ll d$, as a consequence of the data manifold hypothesis. This shows that memorization is directly dependent on the magnitude of the variances when we focus on the single data features.   
Fig.~\ref{fig:tc_comp} shows a comparison between the exact approach for computing $t_c(\vect{x})$ (i.e. using Eqs.~\eqref{eq:zeta}, \eqref{eq:zeta1}) and the small $\alpha$ expansion (i.e. Eq.~\eqref{eq:tcx2}), showing a good qualitative agreement between the two quantities at all values of $\alpha$. The right panel of the same figure also displays a strong dependence of the exactly computed condensation time on the alignment variable $\omega^2(\vect{x})$.   


\subsubsection{Analysis of the empirical Jacobian}\label{app:jacobian}

We can relate this random energy analysis to the spectra of Jacobian eigenvalues using a heuristic argument. In the linear manifold example, the Jacobian of the true score function at $t=0$ is diagonal with eigenvalues equal to $-1/\sigma_k^2$. 
This results in spectral gaps when different sub-spaces have different variances. For a finite value of the effective temperature $t$, the single eigenvalues are equal to $-1/\sigma_k^2 - t^{-1}$. 
After the critical condensation time, the empirical score gives a good approximation of the true score. On the other hand, in the condensation phase the empirical score is dominated by the (quenched) fluctuations in the data distribution. First, we can introduce the participation ratio
\begin{equation}
    Y(t,\vect{x}) = \frac{Z(2/t, \vect{x})}{Z(t, \vect{x})^2}~.
\end{equation}
This thermodynamic quantity can be roughly interpreted as the inverse of the number of energy levels with non-vanishing weights. In the condensation phase, this will be a finite number while it becomes infinite in the high temperature phase. 

In the thermodynamic limit and for $t \leq t_c(\vect{x})$, the participation ratio of our REM model is given by
\begin{equation}
    \mathbb{E}[Y(t,\vect{x})] = 1 - \frac{t}{t_c(\vect{x})}~,
\end{equation}
which implies that the number of data points that contribute to the score function at $\vect{x}$ is 
\begin{equation}
\label{eq:tilda}
    \tilde{N} = e^{\alpha \tilde{d}(t, \vect{x})} = 1/Y(t,\vect{x}) = \frac{1}{1 - t/t_c(\vect{x})}~.
\end{equation}
Note that this number tends to one for $t \to 0^+$, meaning that in the low-time limit the score depends on a single data point. 
Hence in this phase, the score is dominated by approximately $\tilde{N}$ data points leading to the expression
\begin{equation}
    \nabla_{\vect{x}} \log{p_t(\vect{x})} \approx \frac{1}{t\cdot \tilde{N}} \sum_{\mu=1}^{\tilde{N}} \left(\vect{y}^\mu - \vect{x} \right)
\end{equation}
where $\vect{y}^\mu \sim p(\vect{y}^\mu \mid \vect{x}, t) \propto e^{-\frac{1}{2}\vect{y}^T \left( \Lambda^{-1} + t^{-1} I_d \right) \vect{y} + \frac{\vect{x} \cdot \vect{y}}{t}}$. Therefore, the empirical score approximately follows the distribution
\begin{equation}\label{eq:score_normal}
\begin{aligned}
\nabla_{\vect{x}} \log p_t(\vect{x})
\sim &\mathcal{N}\Bigl(
     -M(t)\,\vect{x},\\
    & t^{-1}\bigl(\Lambda^{-1} + t^{-1} I_d\bigr)^{-1}
      \max\!\bigl(0,\, t^{-1} - t_c^{-1}(\vect{x})\bigr)
\Bigr).
\end{aligned}
\end{equation}
where $M(t) = t^{-1}(\Lambda^{-1} + t^{-1} I_d)^{-1} \Lambda^{-1}$ and $\Lambda$ being the diagonal matrix collecting the variances $\sigma_k^2$. 
The $\text{max}$ function in the formula is due to the fact that, for $t > t_c$, an exponentially large number of data points participate in the estimation of the score, which leads to a complete suppression of the variance. On the other hand, the variance of the empirical score estimator diverges for $t \to 0^{+}$. In fact, during condensation, the fluctuations in the random sampling of the data points are not suppressed due to the small number of non-vanishing weights.

We can finally estimate the distribution of the eigenvalues estimated from the empirical Jacobian matrix. Let us set ourselves on $\vect{x} = \vect{0}$, a point belonging to the manifold, and perturb it along the directions of the eigenvectors of $\frac{1}{m}F^{\top} F$. We estimate the elements of the Jacobian of the score function with respect to the orthogonal direction $\vect{e}_j$ using a perturbative approach, i.e. 
\begin{equation}
    J_{ij}(t) \approx \frac{1}{\sqrt{t}} \left(\partial_{x_i} \log p_t\left(\vect{e}_j\cdot\sqrt{t
    }\right) - \partial_{x_i} \log{p_t(\vect{0})}\right).
\end{equation}
Using Eq.~\eqref{eq:score_normal}, we can then write an approximate distribution for the elements of the Jacobian as
\begin{equation}\label{eq:Wij}
\begin{aligned}
J_{ij}(t)
\sim \mathcal{N}\Bigl(
    & -\delta_{ij}\bigl(t+\sigma_i^{2}\bigr)^{-1},\\
    & \frac{\sigma_i^{2}}{t\bigl(t+\sigma_i^{2}\bigr)}
      \bigl[\phi(t,\vect{0}) + \phi(t,\vect{e}_j\sqrt{t})\bigr]
\Bigr).
\end{aligned}
\end{equation}
where we assumed that the fluctuations in $\nabla_{\vect{x}}\log{p_t(\vect{e}_j\cdot\sqrt{t})}$ are independent from the fluctuations in $\nabla_{\vect{x}}\log{p_t(\vect{0})}$ and $\nabla_{\vect{x}}\log{p_t(\vect{e}_j\cdot\sqrt{t} )}$ $\forall k$. In this expression, we introduced the function
\begin{equation}
    \phi(t, \vect{x}) = \max \left(0, t^{-1} - t_c^{-1}(\vect{x}) \right)~.
\end{equation}

We can now recover the singular values of the matrix $J(t)$ as minus the square roots of the eigenvalues of $J(t)^{\top} J(t)$. In general, the matrix $J(t)^{\top} J(t)$ can have a complex spectral distribution. An approximate formula for the singular values of $J(t)$ is
\begin{equation}\label{eq:eigenW}
\begin{aligned}
s_i &\approx
- \Biggl(
    t^{-2}\bigl(1 + t^{-1}\sigma_i^{2}\bigr)^{-2} \\
    &+ t^{-4}\sum_{k=1}^{d}
        \bigl(\sigma_k^{-2}+t^{-1}\bigr)^{-2}
        \bigl[\phi(t,\vect{0})
              + \phi\bigl(t,\vect{e}_i\sqrt{t}\bigr)\bigr]^{2}
\Biggr)^{1/2}.
\end{aligned}
\end{equation}
To obtain this formula, we write $J$ as 
\begin{equation}
    J = A + B
\end{equation}
where $A$ is a diagonal matrix corresponding to the mean of Eq.~\eqref{eq:Wij}, while $B$ corresponds to the variance.
Therefore, $J^\top J$ becomes
\begin{equation}
    J^\top J = A^\top A + A^\top B + B^\top A + B^\top B.
\end{equation}
This expression is dominated by the two symmetric terms, so we can write
\begin{equation}
    J^\top J \approx A^\top A + B^\top B.
\end{equation}
Then, the term $A^\top A = A^2$ is, of course, still diagonal, while the term $B^\top B$ is diagonally dominant. Calling $C = \sum_{ik} B_{ik}B_{ik}$, we can approximate the singular values as $\sqrt{A^2 + C^2}$, obtaining Eq.~\eqref{eq:eigenW}. Note however that the distribution of the spectrum does not concentrate exactly to Eq.~\eqref{eq:eigenW} in the large $N$ limit. Nevertheless, Eq.~\eqref{eq:eigenW} gives an accurate picture of the qualitative behavior, as shown in the main text.

\section{Acknowledgements} 
The work of Bao Pham was funded by the RPI-IBM Future of Computing Research Collaboration (FCRC) program. The results presented here were obtained while Dmitry Krotov was employed by IBM Research. At the time of the present submission Dmitry Krotov is no longer employed by IBM Research. Carlo Lucibello and Enrico Ventura acknowledge the European Union - Next Generation EU fund, component M4.C2, investment 1.1 - CUP J53D23001330001.

\bibliographystyle{plainnat}
\bibliography{bibliography}
\end{document}